\documentclass{article}
\usepackage{amsmath,amssymb}
\DeclareMathOperator{\EX}{\mathbb{E}}
\usepackage[preprint,nonatbib]{neurips_2020}
\usepackage{bbm}
\usepackage{soul}
\usepackage{lipsum}

\usepackage{graphicx}
\usepackage{caption}
\usepackage{amsmath}
\usepackage{xcolor}
\usepackage[ruled,vlined]{algorithm2e}

\DeclareMathOperator*{\argmin}{arg\,min}
\usepackage[utf8]{inputenc}
\usepackage[T1]{fontenc}
\usepackage{hyperref}
\usepackage{url}
\usepackage[ruled,vlined]{algorithm2e}
\usepackage{booktabs}
\usepackage{amsfonts}
\usepackage{nicefrac}
\usepackage{microtype}
\newtheorem{theorem}{Theorem}
\usepackage{color,soul}
\usepackage{bbm}
\usepackage{algpseudocode}
\usepackage{mathtools}

\DeclarePairedDelimiterX{\norm}[1]{\lVert}{\rVert}{#1}
\title{Vulnerability Under Adversarial Machine Learning:  Bias or Variance? }

\author{
  \normalfont{Hossein Aboutalebi\textsuperscript{1}, Mohammad Javad Shafiee\textsuperscript{1}}\\
Michelle Karg\textsuperscript{2}, Christian Scharfenberger\textsuperscript{2}\\
 Alexander Wong\textsuperscript{1}\\
 \textsuperscript{1}Waterloo AI Institute, University of Waterloo, Waterloo, Ontario, Canada\\
  \textsuperscript{2}ADC Automotive Distance Control Systems GmbH, Continental, Germany \\
  \textsuperscript{1}\{haboutal, mjshafiee, a28wong\}@uwaterloo.ca \\
  \textsuperscript{2}\{michelle.karg, christian.scharfenberger\}@continental-corporation.com
}

\begin{document}

	 \maketitle

\begin{abstract}
   Prior studies have unveiled the vulnerability of the deep neural networks in the context of adversarial machine learning, leading to great recent attention into this area.  One interesting question that has yet to be fully explored is the bias-variance relationship of adversarial machine learning, which can potentially provide deeper insights into this behaviour. The notion of bias and variance is one of the main approaches to analyze and evaluate the generalization and reliability of a machine learning model. Although it has been extensively used in other machine learning models, it is not well explored in the field of deep learning and it is even less explored in the area of adversarial machine learning.
   In this study, we investigate the effect of adversarial machine learning on the bias and variance of a trained deep neural network and analyze how adversarial perturbations can affect the generalization of a network. We derive the bias-variance trade-off for both classification and regression applications based on two main loss functions: (i) mean squared error (MSE), and (ii) cross-entropy.  Furthermore, we perform quantitative analysis with both simulated and real data to empirically evaluate consistency with the derived bias-variance tradeoffs. Our analysis sheds light on why the deep neural networks have poor performance under adversarial perturbation from a bias-variance point of view and how this type of perturbation would change the performance of a network. Moreover, given these new theoretical findings, we introduce a new adversarial machine learning algorithm with lower computational complexity than well-known adversarial machine learning strategies (e.g., PGD) while providing a high success rate in fooling deep neural networks in lower perturbation magnitudes.
\end{abstract}
\section{Introduction}
\vspace{-0.25cm}
Despite of the impressive achievements of deep learning  over the past decade in different fields such as computer vision~\cite{he2016deep,krizhevsky2012imagenet, lecun2010convolutional,  redmon2016you}, machine translation~\cite{vaswani2017attention, wu2016google},  and medicine~\cite{cruz2017accurate,de2018clinically}, their vulnerability against adversarial machine learning  brings different concerns regarding their robustness.

A perturbation $\epsilon$ in a specific direction to the input causes the model to incorrectly classify the input sample which can be preformed in both classification~\cite{moosavi2016deepfool, szegedy2013intriguing}  or regression problems~\cite{alfeld2016data,tong2018adversarial}. The perturbation $\epsilon$ should be imperceptible  by a human eye and as such, the norm of $\epsilon$ is bounded when a new perturbation is generated. \mbox{Szegedy {\it et al.}} introduced this drawback for deep neural networks in their seminal paper~\cite{szegedy2013intriguing}. They observed that the state-of-the-art deep neural networks act poorly  with high confidence when an imperceptible non-random perturbation is added to the input image. They attributed this poor behaviour to the potential blind spots in the training of deep neural networks.  Goodfellow {\it et al.}~\cite{goodfellow2014explaining} argued this poor performance of deep neural networks on adversarial examples is due to their linear behavior in high-dimensional spaces.
Since then, there have been several studies introducing different approaches to generate adversarial perturbation and fool the deep neural networks~\cite{goodfellow2014explaining, huang2015learning, moosavi2016deepfool,nguyen2015deep}. However, Madry {\it et al.}~\cite{madry2017towards}  proposed  a multi-step attack called the projected gradient descent (PGD) algorithm which generalizes the prior first-order adversarial machine learning algorithms and is able to produce adversarial examples that are harder to learn and to defeat.

Parallel to introducing new adversarial machine learning  algorithms, there has been a line of research  focusing on building defense mechanism against adversarial machine learning methods. Adversarial training initially proposed by Goodfellow {\it et al.}~\cite{goodfellow2014explaining} is one of the pioneer works which augments the training set with adversarial examples to improve the network resilience against adversarial machine learning algorithms. Kurakin \textit{et al.}~\cite{kurakin2016adversarial} extended this approach where they demonstrated the success of this method  for Inception v3 model~\cite{szegedy2016rethinking} trained on the ImageNet dataset~\cite{russakovsky2015imagenet}. Furthermore, they illustrated that one-step attacks like FGSM have a better transferability properties between models as a black-box attack where the attacker does not have access to the parameters of the model. \mbox{Madry \textit{et al.}~\cite{madry2017towards}} demonstrated that the higher the capacity of the deep neural network model, the more resilient it is against adversarial attacks. The adversarial training method is further extended by Liu \textit{et al.}~\cite{liu2018adv} (so-called Adv-BNN) where Bayesian techniques benefit the adversarial training method to improve the robustness of the model. Besides adversarial training, other approaches such as defensive distillation~\cite{papernot2016distillation} generates a new set of soft training labels by replacing the Softmax outputs of a neural network with a smoothed values. A new network with a similar architecture  is then trained using the new training set. Although this approach improves the robustness of the model against some simpler adversarial machine learning methods, it behaves poorly against more recent attacks like CW adversarial machine learning~\cite{carlini2017towards}.
While the main focus of the adversarial machine learning has been for classification tasks, several approaches have been proposed to apply them on regression problems as well~\cite{balda2019perturbation,ghafouri2018adversarial, nguyen2018adversarial}. Sensory data  is one of the main field of research for adversarial machine learning algorithms with regression focus. For example, Ghafouri {\it et al.}~\cite{ghafouri2018adversarial} studied the detection of adversarial attacks against cyber-physical systems (CPS)  capable of manipulating sensor readings.

Despite a rich literature developed in the field of adversarial machine learning, there has not been  enough theoretical studies on why neural networks are vulnerable in facing inputs perturbed with adversarial perturbations. Zhang \textit{et al.}~\cite{zhang2019theoretically} analyzed the robustness and accuracy trade-off in deep neural networks. They illustrated that the side-effect of making a model robust against adversarial machine learning via adversarial training is a drop in the accuracy of the model. They provided an upper bound on the gap between the robustness and accuracy. As such, model generalization is an important drawbacks of these techniques.
Bias and Variance are one of the long-standing and well-known procedure to analyze the generalization and reliability of machine learning models. The seminal work by Geman \textit{et al.}~\cite{geman1992neural} showed that  while a model's variance increases, the model's bias decreases monotonically with the increase in the model complexity. They derived a \mbox{well-formed} decomposition of the bias and variance of the loss function for the regression learning task. Domingos~\cite{domingos2000unified} extended the bias-variance decomposition to a more general loss functions such as cross-entropy. However, obtaining a well-formed decomposition for a cross-entropy loss function is more difficult than the case of mean squared error (MSE) loss function~\cite{domingos2000unified}. Recently, Neal \textit{et al.}~\cite{neal2018modern} challenged the earlier findings of  Geman \textit{et al.}~\cite{geman1992neural}  by demonstrating that the variance for  deep neural networks does not necessarily increase as the network width increases. Followed by that, a more recent work by Yang~\textit{et~al.}~\cite{yang2020rethinking} closed the gap between these two studies by showing that the variance has a bell-shaped behaviour with the model complexity. In other words, as the width of the network increases, the variance first starts to increase and then decreases. They backed their claim with both theoretical analysis and empirical examples.

Although bias-variance trade-off has been used to justify some aspects of  deep neural networks in previous studies, to the best of authors' knowledge, the theoretical analysis around the impact of the adversarial machine learning algorithms on the bias and variance of a deep neural network has not been well explored.
In this paper, we aim to study the effect of adversarial machine learning on the bias and variance of a deep neural network.
 Here, a new decomposition of the loss function in a deep neural network is derived in terms of its bias and variance for both the regression and classification tasks when the input sample is perturbed by an adversarial machine learning algorithm.
 Our main contributions are as follow:
 \vspace{-0.5cm}
 \begin{itemize}
     \item The bias and variance of a deep neural network facing adversarial perturbations is decomposed for both MSE and cross-entropy loss functions.
     \item The new derivations illustrate what should be the behavior of the adversarial machine learning method to enforce the maximimum changes in the network's loss function and maximize the success rate.
     \item Extensive experimental results validate the new theoretical findings in the network's bias and variance theorem for both MSE and cross-entropy loss functions.
     \item A new adversarial machine learning method (so-called BV adversarial attack) is proposed which is capable of fooling deep neural networks with comparable results with the state-of-the-art algorithms but with higher efficiency and less computational complexity.
 \end{itemize}
 The proposed theorems illustrate that an adversarial machine learning algorithm  can be designed in such a way that attacks a model by only changing its behaviour in terms of either  bias or variance. This finding is also observed  experimentally in the reported result of Figure~\ref{fig:fgsm-simulated} (c)  for the case of MSE for a regression problem. As such, the proposed theorems suggest that it is possible to design more powerful adversarial machine learning algorithms which are much harder to be detected and resolved. One interesting idea would be to design adversarial machine learning methods which only change the model variance and only make the model unstable in specific cases and situations. As a result, they might be very hard to identify as there is not any significant change in the model's bias which make them more disastrous.

 \vspace{-0.35cm}
 \section{Methodology}
 \vspace{-0.25cm}
 In this section we illustrate the effect of adversarial machine learning algorithms on the model's bias and variance and derive how the perturbation can change the behaviour of a model by studying its bias and variance.
 Here we aim to study the bias-variance trade-off in deep neural networks based on two well-know loss functions, MSE loss and cross-entropy loss.
\vspace{-0.35cm}
\subsection{Notation}

$\norm{x}$ denotes a generic norm function. Notations $\norm{x}_2$, $\norm{x}_\infty$ refers to the $l_2$ and $l_{\infty}$ norms respectively. A set is denoted with capital letters such as ~$\mathcal{X}, \mathcal{Y}$ while vectors are denoted by small letters such as $x$ and $y$. The training set is denoted with $\mathcal{D}$ and the target function by $f: \mathcal{X} \rightarrow \mathcal{Y}$. In case of regression learning tasks, the set $\mathcal{Y}$ is a continuous one dimensional space while in the classification task it contains discrete values. In our setting, a prediction model is denoted by $\hat{f}$ which is an estimation of the ground truth function $f$ over the training set $\mathcal{D}$. In order to consider adversarial perturbation, we denote the perturbation added to each data sample $x$ by an adversary with the vector $\beta(x)$. $\beta(x)$ is not generated naturally and is designed specifically for each data sample and usually has the property $l_{\infty}\big(\beta(x)\big)< \delta$. Throughout our analysis, whenever we use the notation $\nabla f(x)$, it is the gradient of the function $f$ with respect to $x$.

\subsection{Case I: Regression with MSE Loss}
Assume the goal is to estimate the target function $f: \mathcal{X} \rightarrow \mathcal{Y}$. Each element $x \in \mathcal{X}$ has dimension $|x|=d$. Given the training data, \mbox{$\mathcal{D} =\big\{(x_1,y_1),..., (x_m,y_m)\big\}$}, a learner produces a prediction model $\hat{f}(x)$. As such, the configuration of the parameters in $\hat{f}(x)$ is dependent on  the training data $\mathcal{D}$. Let us also assume  the training data $\mathcal{D}$ is accompanied with a  natural noise $\gamma$ such that:
\begin{align}
\begin{small}
\label{zeroo}
    y_i=f(x_i)+\gamma
\end{small}
\end{align}
where $1\leq i\leq m$ with $m$ total number of data samples in $\mathcal{D}$, and $\gamma$ is a random variable where $\EX[\gamma]=0$, and  $\EX[\gamma^2]=\sigma^2_\gamma$. It is worth to note that, we keep this assumption mainly for the regression task and we will drop it for the classification problems with cross-entropy loss function for simplicity.

Geman \textit{et al.}~{\cite{geman1992neural}} decomposed a MSE loss function in terms of its bias and variance of a prediction model by Theorem~\ref{th1}.
\begin{theorem} \label{th1}For a prediction model $\hat{f}(x)$ trained on the training data $\mathcal{D}$ to estimate the target function $f(x)$ with MSE loss function, the bias variance trade-off is \cite{geman1992neural}:
 \begin{small}
 	\begin{align}\label{eq2}
 	\EX_{x,\mathcal{D},\gamma}\left[(y-\hat{f}(x))^2\right]&=\EX_{x,\mathcal{D}}\Big[\big(\EX_\mathcal{D}\big[\hat{f}(x)\big]-f(x)\big)^2\Big]+
 	\EX_{x,\mathcal{D}}\Big[\big(\hat{f}(x)-\EX_\mathcal{D}\big[\hat{f}(x)\big]\big)^2\Big]+\sigma^2_\gamma\nonumber\\
 	&=Bias[\hat{f}]+Var[\hat{f}]+Var[\gamma].
 	\end{align}
 \end{small}
 \end{theorem}

 The $Var[\gamma]$ is the intrinsic noise of the system.
 	Given~\eqref{eq2}, it is possible to break down and decouple the effect of different factors on model performance based on  the bias, variance and intrinsic noise in the model.
 	However, \eqref{eq2} does not take  the effect of adversarial perturbation into account. The perturbation $\beta(x)$ added to each data sample $x$ during the test time aims to increase the loss value of the model. It is assumed that $f(x) = f(x+\beta(x))$, this assumption is to make sure the added perturbation magnitude is reasonable and follows the imperceptibility of the adversarial perturbation.
 	
 	This perturbation can have a great impact on the final loss which is significantly different from {\eqref{eq2}}. Following, we propose a new theorem to account for the adversarial perturbation in deriving bias and variance of a model.
 	
 \begin{theorem}\label{th2} Assume $\bar{f}(x)=\EX_\mathcal{D}[\hat{f}(x)]$ and the target function is $f(x)$. The bias-variance trade-off for MSE loss function with a prediction model $\hat{f}(x)$ trained on dataset $\mathcal{D}$ with noise $\gamma$ in the presence of adversarial perturbation $\beta(x)$ via the adversarial algorithm  is:
 	\begin{small}
 \begin{align} \label{eq7}
 &    \EX_{x,\mathcal{D},\gamma}\left[(y-\hat{f}(x+\beta(x)))^2\right]\approx\EX_{x,\mathcal{D}}[(f(x)-\bar{f}(x)-c_x)^2]+Var[\gamma]+Var[\hat{f}]+\EX_{x,\mathcal{D}}[c_x'] \\
&\text{where,   } \;\;\;c_x =\nabla\bar{f}(x)^T\beta(x)  \;\;\text{and,} \;\; c_x'= 2\Big(\hat{f}(x)-\bar{f}(x)\Big) \Big(\Big(\nabla\hat{f}(x)-\nabla\bar{f}(x)\Big)^T\beta(x)\Big)
\label{eq:cx_tem}
\end{align}
	\end{small}
\end{theorem}
\textbf{Proof:} Given the adversarial perturbation $\beta(x)$ with the condition $l_{\infty}\big(\beta(x)\big)< \delta$ during the test time\footnote{The detailed derivation can be found in supplementary material.},
	\begin{small}
\begin{align}\label{eq3}
	&\EX_{x,\mathcal{D},\gamma}\left[(y-\hat{f}(x+\beta(x)))^2\right]=\EX_{x,\mathcal{D},\gamma}\left[(f(x)-\hat{f}(x+\beta(x))+\gamma)^2\right]\nonumber
	\\&=\EX_{x,\mathcal{D}}\left[(f(x)-\bar{f}(x+\beta(x)))^2\right]+\EX_{x,\mathcal{D}}\left[(\hat{f}(x+\beta(x))-\bar{f}(x+\beta(x)))^2	\right]+\sigma^2_\gamma
	\nonumber
	\\&\approx \EX_{x,\mathcal{D}}\left[\underbrace{\left(f(x)-\bar{f}(x)-\nabla\bar{f}(x)^T\beta(x)\right)^2}_{Bias}\right]+\EX_{x,\mathcal{D}}\left[\underbrace{(\hat{f}(x+\beta(x))-\bar{f}(x+\beta(x)))^2}_{Variance}	\right]+\sigma^2_\gamma
\end{align}
	\end{small}
As seen in~\eqref{eq3}, $c_x =\nabla\bar{f}(x)^T\cdot\beta(x) $ is the new term added to the bias of the model because of the adversarial perturbation.  Next we expand the variance term in~\eqref{eq3} to illustrate how the adversarial perturbation affects the variance of the model:

	\begin{small}
\begin{align}\label{var1}
    \EX_{x,D}&\left[(\hat{f}(x+\beta(x))-\bar{f}(x+\beta(x)))^2
    \right]\approx\EX_{x,D}\left[\left(\hat{f}(x)-\bar{f}(x)\right)^2 +2(\hat{f}(x)-\bar{f}(x))(\nabla\hat{f}(x)-\nabla\bar{f}(x))^T\beta(x) \right]\nonumber\\
    &=Var[\hat{f}]+\EX_{x,D}\left[2(\hat{f}(x)-\bar{f}(x))(\nabla\hat{f}(x)-\nabla\bar{f}(x))^T\beta(x)\right]
\end{align}
	\end{small}

\hfill$\blacksquare$\\
As illustrated, one important conclusion from the Theorem~{\ref{th2}} is that the adversarial machine learning increases the bias term (i.e., $c_x$ in \eqref{eq:cx_tem}) and it can also increase the variance of the model by $c_x'$ in~\eqref{eq:cx_tem}. Experimental results verify these findings as well. The following corollaries are the direct results of theorem {\ref{th2}}.

\textbf{Corollary I: }The maximum expected increase in the bias of a model with MSE loss function for a regression task  is when the  adversarial perturbation is added in the direction $-\nabla\bar{f}(x)$.

\textbf{Corollary II: }The maximum expected increase in the variance of a model trained for a regression task with MSE loss function is when the adversarial perturbation is added in the direction of $(\hat{f}(x)-\bar{f}(x))(\nabla\hat{f}(x)-\nabla\bar{f}(x))$.

\subsection{Case II: Classification with cross-entropy Loss}

The notion of bias and variance can be analyzed for the classification models trained with cross-entropy loss as well. To this end, followed by the work done in~\cite{pfau2013generalized,yang2020rethinking} let $c$ be the number of classes for classification and $\hat{\pi}_\mathcal{D}(x) \in [0,1]^c$ be the output of a neural network trained on the training set $\mathcal{D}$.  This function measures the confidence values over classes. Let $\pi(x) \in [0,1]^c$ be a one-hot vector encoding ground truth label that we wish to estimate via $\hat{\pi}$. Then cross-entropy loss can be formulated as:
	\begin{small}
\begin{align}\label{cross0}
    L(\pi,\hat{\pi})=-\EX\limits_{x,\mathcal{\mathcal{D}}}\left[\sum_{i=1}^c \big(\pi_i(x)\log \hat{\pi}_i(x)\big)\right]
\end{align}
	\end{small}
\noindent where  $\pi_i(x)$ refers to $i$th component of the output vector $\pi(x)$. As explained in \cite{pfau2013generalized}, the loss function in~\eqref{cross0} can be decomposed:
	\begin{small}
\begin{align} \label{cross11}
    L(\pi,\hat{\pi})=\EX\limits_{x,\mathcal{\mathcal{D}}}\big[D_{KL}\left(\pi(x)||\pi^*(x)\right)+D_{KL}\left(\pi^*(x)||\hat{\pi}(x)\right)\big]
\end{align}
	\end{small}
where $\pi^*(x)=\argmin_z \EX\limits_{\mathcal{D}}\left[D_{KL}(z||\hat{\pi})\right]$.  $\pi^*(x)$ is the prediction model that has the minimum expected KL-Divergence from the possible prediction models space. In other words, we can consider it as the mean of the prediction model $\hat{\pi}$ which is defined in terms of KL-Divergence. This perspective   which is further elaborated in Domingos's seminal work~\cite{domingos2000unified}  is somehow different from the mean defined in the previous section for the regression task because of the cross-entropy loss function form. As a result, it is possible to consider $D_{KL}(\pi(x)||\pi^*(x))$ as the factor which drives the bias  and  $D_{KL}(\pi^*(x)||\hat{\pi}(x))$ as the one deriving the variance in the model. However to account for the adversarial perturbation instead of input $x$, the function ${\pi}^*(\cdot)$ needs to be calculated for $x+\beta(x)$. This leads to Theorem~\ref{th3} which illustrates the behavior of a model trained based on cross-entropy loss in the presence of adversarial perturbation $\beta(x)$.  It is assumed that the target function $\pi(x)$ is constant on a small blob around $x$, and $\beta(x)$ magnitude does not exceed the limits of that blob. In other words, $\pi(x)=\pi(x+\beta(x))$.
\begin{theorem}\label{th3} Assume for input $x$, the ground truth class is $t_x$. For a cross-entropy loss function, the bias-variance tradeoff of a prediction model $\hat{\pi}(x)$ with training data $\mathcal{D}$ for a target function $\pi(x)$ in the presence of adversarial algorithm injecting perturbation $\beta(x)$ to the system is:
	\begin{small}
\begin{align} \label{cross44}
        L(\pi,\hat{\pi})=\EX\limits_{x,\mathcal{D}}\big[D_{KL}(\pi(x)||{\pi}^*(x))+D_{KL}({\pi}^*(x)||\hat{\pi}(x))\big]+\EX\limits_{x}\left[c_x\right] +\EX\limits_{x,\mathcal{D}}\left[c'_x\right]
    \end{align}
\end{small}
    where,
\begin{small}
    \begin{align}
    &c_x=-\EX\limits_{x}\left[\big{(}\nabla_x\log{\pi}^*_{t_x}(x)\big{)}^T\beta(x)\right]
    &c'_x=-\EX\limits_{x,\mathcal{D}}\left[\sum_{i=1}^c\big{(}\nabla_x{\pi}^*_i(x)\log\frac{\hat{\pi}_i(x)}{{\pi}^*_i(x)}\big{)}^T\beta(x)\right]
    \label{eq:cx}
    \end{align}
\end{small}
\end{theorem}
\textbf{Proof: }The proof can be found in the supplementary material.

This derivation is aligned with finding in~\cite{pfau2013generalized,yang2020rethinking}, where the bias variance decomposition for cross-entropy loss function is in the form of KL-Divergence. The proposed theorem leads to the following corollaries,\\
\textbf{Corollary I: }The maximum expected increase in the bias of a deep neural network trained with cross-entropy loss is when the adversarial perturbation is in the direction of $c_x$ in~\eqref{eq:cx}.\\
\textbf{Corollary II:} The maximum expected increase in the variance of a deep neural network trained with cross-entropy loss is when the adversarial perturbation is in the direction of $c'_x$ in~\eqref{eq:cx}.\\

\vspace{-0.25cm}
\subsection{BV Adversarial Machine Learning}
\vspace{-0.25cm}

Despite the fact that Theorem~\ref{th3} provides a solid bias-variance decomposition and gives directions of $c_x, c'_x$ for maximum change on bias and variance of the system respectively, it is not computationally feasible to calculate $\pi^*(x)$. As a result, in this section the cross-entropy loss function is analyzed in a different way. This new finding leads to a novel adversarial machine learning algorithm which can be used to fool models with less computational complexity and a reasonable success rate.

Given a classification problem with c classes, there is a set of score functions $\{\hat{f}_1(x),\hat{f}_2(x), ..., \hat{f}_c(x)\}$  which measure the possibility of assigning  input $x$ to each class $1,2, ..., c$  respectively. These scores are non-negative and encode the confidence of each function to assign the sample to the their corresponding class labels.

For an input $x \in \mathcal{X}$, let us assume there is a target function $f(x)$ which maps the input $x$ to the set $\{1,2,...,c\}$. Define $X_i=\{x| x\in \mathcal{X} \wedge f(x)=i\}$ where $i \in \{1,...,c\}$. Given a cross-entropy loss function, the error of the prediction functions $\hat{f}_i(x)$ for $i \in \{1,...,c\}$ of the target function $f(x)$ is:
	\begin{small}
\begin{align}\label{cross1}
    \EX_{x,\mathcal{D}}\left[\sum_{i=1}^c\left(-\log(\frac{\hat{f}_i(x)}{\hat{f}_1(x)+...+\hat{f}_c(x)})\right)\mathbbmss{1}_{f(x)=i}\right] =\sum_{i=1}^c\EX_{x\in X_i,\mathcal{D}}\left[-\log\big(\hat{f}_i(x)\big)+\log\big(\hat{f}_1(x)+...+\hat{f}_c(x)\big)\right].
\end{align}
	\end{small}
As such, it is possible to decompose the cross-entropy loss function based on its bias and variance as follow,
\begin{theorem} \label{th4} The decomposition of the cross-entropy loss function for the prediction model  constructed by  the set of functions $\{\hat{f}_1(x), ..., \hat{f}_c(x)\}$ with training data $\mathcal{D}$ and the target function $f(x)$ with $c$ classes in the presence of adversarial  perturbation of $\beta(x)$ is:
	\begin{small}
 \begin{align} \label{eq7}
     \EX_{x,\mathcal{D}}\left[\sum_{i=1}^c\left(-\log(\frac{\hat{f}_i(x+\beta(x))}{\hat{f}_1(x+\beta(x))+...+\hat{f}_c(x+\beta(x))})\right)\mathbbmss{1}_{f(x)=i}\right]
     \approx \sum_{i=1}^c\EX_{x \in X_i,\mathcal{D}}\left[-\log(\hat{f}_i(x))+\log(\hat{f}_1(x)+...+\hat{f}_c(x))+c_i(x)\right]
 \end{align}
	\end{small}
\begin{small}
\begin{align}
\text{Where,  }\;\;\ c_i(x) =-{\nabla} log\left(\frac{\hat{f}_i(x)}{\hat{f}_1(x)+...+\hat{f}_c(x)}\right)^T\cdot\beta(x)  \;\;\;\text{    s.t.   }\;\;\;\; X_i=\Big\{x| x\in X \wedge f(x)=i\Big\}.
\end{align}
\end{small}
\end{theorem}

\textbf{Proof: } The proof is included in the supplementary material.

\begin{algorithm}[t]

	\KwData{$\Big\{\big(x,y(x)\big)| x \in D\Big\}$ with $c$ distinct classes}
	\KwResult{$\hat{x}$ \Comment{Perturbed image $x$. }}
	\textbf{Input}: $\hat{f_i} \;\;\;\; i \in \{1,...,c\}$, \Comment{The prediction model scores for all classes.}\\
	~~~~~~~~~~~~$\epsilon$, \Comment{The magnitude of perturbation.}\\
	~~~~~~~~~~~~$x$, \Comment{The input image.}\\
	~~~~~~~~~~~~$y$,\Comment{ The ground truth label.}\\
	\textbf{Begin}\\
		\begin{small}
	~~~~~$S=[- {\nabla}_x \log\left(\frac{\hat{f}_1(x)}{\hat{f}_1(x)+...+\hat{f}_c(x)}\right),..., -{\nabla_x} \log\left(\frac{\hat{f}_c(x)}{\hat{f}_1(x)+...+\hat{f}_c(x)}\right)\Big]$\\
	~~~~~$V = \Big[l_i|l_i = \mathbbmss{1}_{y = i}\; , \;i=1,\ldots,c\Big]$ \Comment {The one-hot vector of the ground truth label.}\\
	~~~~~$\hat{x}= x+\epsilon~S^T V$  \\
		\end{small}
	~~~~~ \textbf{Return}~ $\hat{x}$\\
	\textbf{End}
	\caption{BV Attack for Multi-class classification with $c$ classes }
	\label{alg:bv}
\end{algorithm}

Given the derivation in~{\eqref{eq7}}, the maximum increase in the loss function for class $i$ is reached when the attacker attacks in the direction of $-{\nabla} log\left(\frac{\hat{f}_i(x)}{\hat{f}_1(x)+...+\hat{f}_c(x)}\right)$ for $x\in X_i$.
As a result,
while~{\eqref{eq7}} illustrates how the adversarial perturbation can affect the model's loss decomposition, it also provides the new terms $c_i(x)$ in the equation which gives the direction to maximize its attack on the prediction model's loss.

Motivated by that, using the direction vector $c_i(x)$ in Theorem {\ref{th4}}, we can derive a new attack on the prediction model. The procedure to perform the proposed attack (so-called BV attack) is summarized in Algorithm~\ref{alg:bv} for multi-class classification task with $c$ classes. It is assumed that  each $\hat{f}_i$  (i.e., the confidence function for class $i$) outputs a non-negative value. This assumption is aligned with the current design of deep neural networks for classifications, as the final output is usually passed through the Softmax layer to be normalized.

The model's gradient associated with the ground truth label of the input $x$ is added to the input; and the magnitude of the perturbation  is controlled by the value of $\epsilon$. However it is worth to note that since the intrinsic properties of the proposed BV adversarial perturbation is different compared to other adversarial machine learning method, it is not possible to compare the $\epsilon$ in the proposed BV method and other adversarial machine learning algorithms directly. As such, the competing methods are compared based on the perturbation and the amount of noise they add to the input sample.
\vspace{-0.5cm}
\section{Experimental Results \& Discussion}
\vspace{-0.25cm}
In this section, we examine the proposed theorems experimentally and illustrate how a deep neural network behaves facing an adversarial machine learning algorithm based on its bias and variance.
To this end, we evaluate the theorem via different neural network architectures on both real datasets including  CIFAR-10 and CIFAR-100~\cite{krizhevsky2009learning} and simulated data for both classification and regression tasks.
\vspace{-0.5cm}
\subsection{Simulated Data}
\begin{figure*}
\vspace{-1.5cm}
\hspace*{-1.5cm}
\setlength{\tabcolsep}{0.01cm}
\begin{tabular}{ccc}
        \includegraphics[width=0.37\textwidth]{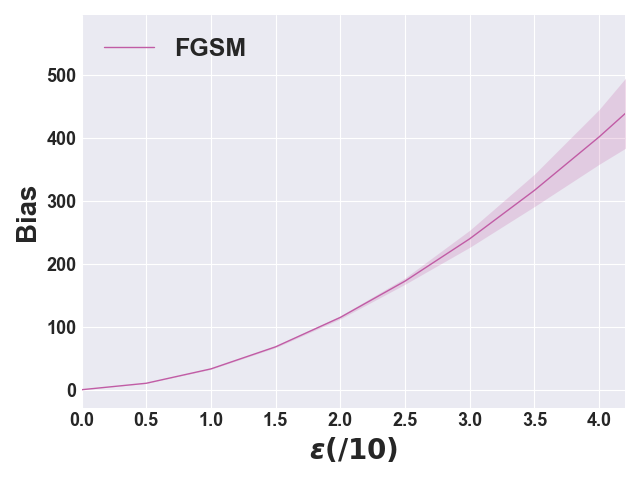}&
		\includegraphics[width=0.37\textwidth]{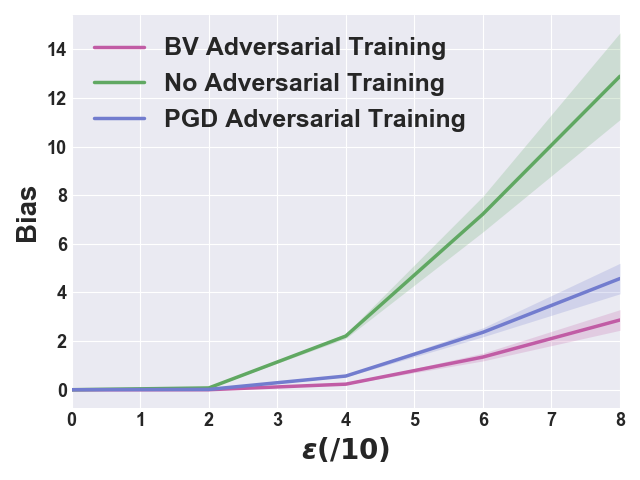}&
			\includegraphics[width=0.37\textwidth]{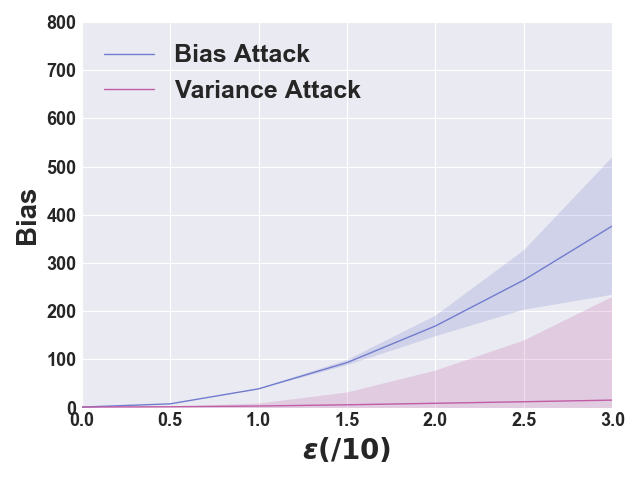}\\
		(a)  Regression--FGSM  Attack&  (b) Classification--FGSM Attack & (c) Regression--Bias-Variance Attack
\end{tabular}
\caption{(a)--(b) The effect of FGSM adversarial machine learning on bias and variance for models via the simulated data in both regression and classification tasks. (c) The BV adversarial machine learning algorithm regression task when bias or variance perturbations derived in Theorem \ref{th2} is being used to perturb the input sample.}
\vspace{-0.5cm}
\label{fig:fgsm-simulated}
\end{figure*}

\vspace{-0.25cm}
As the first experiment, the proposed theorems are evaluated based on the simulated data.
For the regression problem, samples $(x,y)$, from a linear target function with natural noise of $\gamma$ (generated from a Uniform distribution) are generated. Each sample $x \in \mathbbmss{R}^2$ should be mapped to $y\in \mathbbmss{R}$. A simple feed forward neural network with one hidden layer of $100$ units is used to   learn the target function. 5 experiments with different  seeds are performed to measure the bias and variance of the model given the perturbed samples by FGSM adversarial attacks (Figure~\ref{fig:fgsm-simulated} (a)).

A similar experiment is performed for classification task as well. The simulated samples for two classes \mbox{$(x\in \mathbbmss{R}^{50}, y \in \{0,1\})$} are generated from Normal distributions $\mathcal{N}(0,1)$ and $\mathcal{N}(10,1)$ to have enough separability between two class labels. The training samples are modeled with a feed forward network logistic regression model with two hidden layers with sizes $[50,100]$. This trained model is attacked by FGSM.
Figure~\ref{fig:fgsm-simulated} (b) demonstrates the experimental results. As seen, by increasing the attack (i.e., increasing the $\epsilon$) the bias and variance of the models are increased which is aligned by the proposed theorems. The experiment is performed when model is trained without any adversarial training and when it is trained by PGD or BV adversarial training techniques.

As the last experiment, the findings in Theorem~\ref{th2} is analyzed experimentally. To do so, the regression model trained in the first experiment is attacked by the perturbations derived in Theorem~\ref{th2}. As seen in Figure~\ref{fig:fgsm-simulated} (c), perturbing the input sample by $-\nabla\bar{f}(x)$ (i.e., Bias attack) enforces the maximum bias change in the model while using $(\hat{f}(x)-\bar{f}(x))(\nabla\hat{f}(x)-\nabla\bar{f}(x))$ (i.e., Variance attack) would change the model variance with a minimum change in the model's bias. This experiment validates our finding in Theorem~\ref{th2} experimentally.

\vspace{-0.5cm}
\subsection{Real Data \& BV Adversarial Attack Algorithm}
\vspace{-0.25cm}
In this section, we study the effect of adversarial machine learning algorithms on real datasets. To this end, we take advantage of adversarial training techniques during the training to improve the robustness of the examined models. Four different deep neural networks including \mbox{ResNet-18}~\cite{he2016deep}, ResNet-34, ResNet-50\footnote{Here we report the results for two models and the complete experimental results can be found in the supplementary material.} and MobileNetV2~\cite{sandler2018mobilenetv2} is used to conduct the experimental result for \mbox{CIFAR-10} and CIFAR-100 datasets. Each experiment is repeated by 5 different seeds to be able to calculate the bias and variance for the models, properly.

Figure~\ref{fig:acc-cifar10} shows the experimental results. Columns (a) demonstrates the effect of different adversarial machine learning methods on the accuracy of four examined adversarial machine learning algorithms when they are trained via a regular training approach\footnote{The behaviour of loss functions for the examined models are included in the supplementary material.}. Since the intrinsic behaviour of $\epsilon$ in the evaluated adversarial machine learning methods are different, the examined models' accuracy are compared based on the perturbation level of the perturbed input samples which is measured by the MSE difference of the ground truth image and the perturbed one.  As seen, by increasing the perturbation level the model's accuracy drops (i.e., change in the model bias), the model variance is increased as well. One interesting observation is that while the variance of the loss value keeps increasing with the perturbation level, the variance of the accuracy starts to decrease when the perturbation passes a threshold. That is due to the fact that extreme perturbation level causes the  model to have almost zero accuracy which is also visible in Figure~{\ref{fig:acc-cifar10}} and as such there is not significant changes in the variance. This result is consistent with the bias-variance tradeoff presented in Theorem~\ref{th3} and~\ref{th4}. Columns (b)--(d) show  the same experiment when an adversarial training technique is used during training of the model. The results illustrate that using the adversarial training technique improves the robustness of the model and leads to lower bias  while facing higher perturbation compared to the a regular training approach. Furthermore, the result shows that adversarial training technique can benefit the model and reduces the model variance given different perturbation levels.

\begin{figure*}
\vspace{-0.35cm}
\hspace*{-1.5in}
\footnotesize
\setlength{\tabcolsep}{0.01cm}
\centering
        \begin{tabular}{cccc}
        \includegraphics[width=0.37\textwidth]{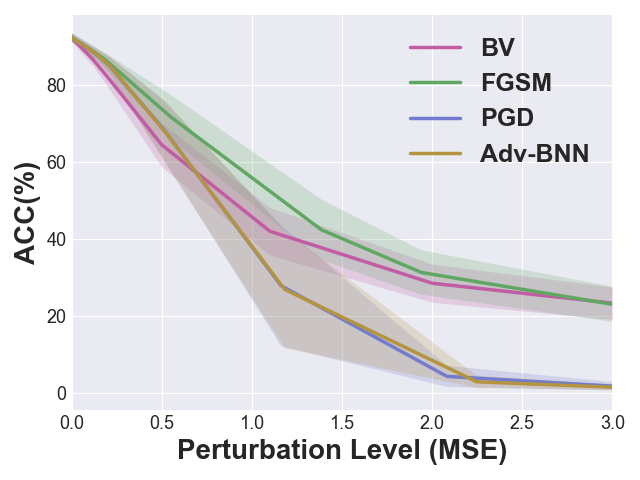}&
		\includegraphics[width=0.37\textwidth]{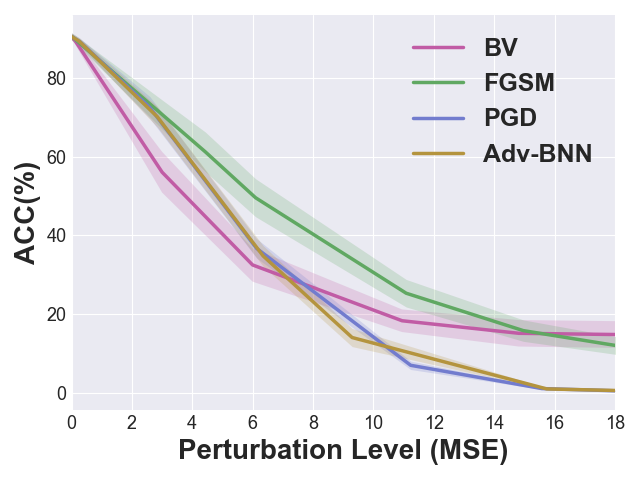}&
		\includegraphics[width=0.37\textwidth]{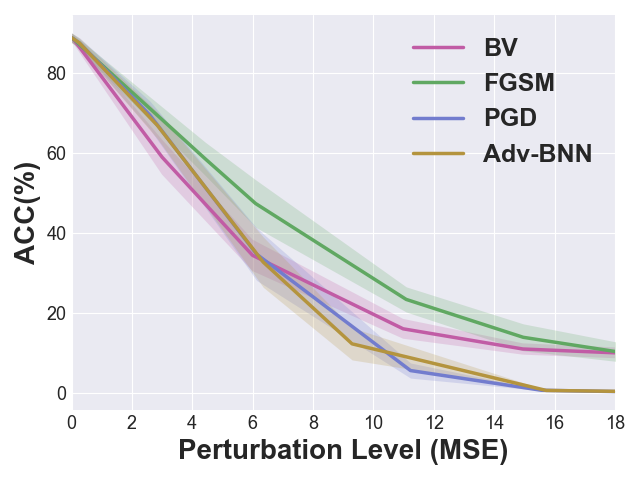}&
		\includegraphics[width=0.37\textwidth]{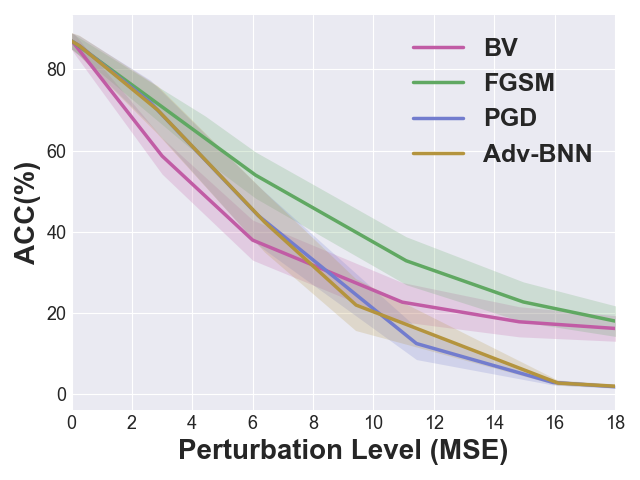}\\
		(a)  ResNet18--No Adversarial Training & (b) ResNet18--FGSM Adv-Training&  (c) ResNet18--BV Adv-Training& (d) ResNet18--PGD Adv-Training \\
		\includegraphics[width=0.37\textwidth]{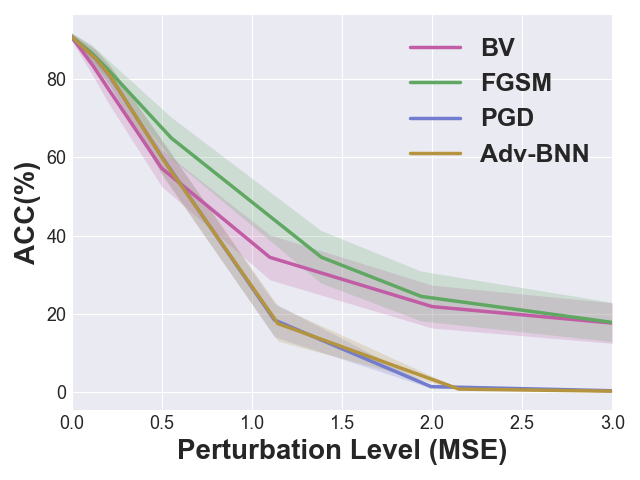}&
		\includegraphics[width=0.37\textwidth]{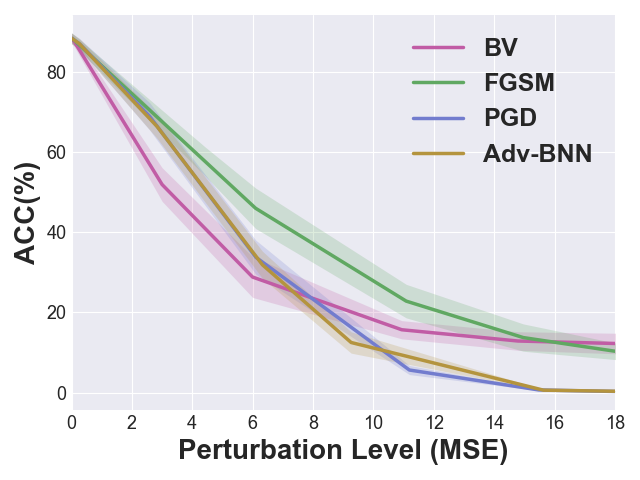}&
		\includegraphics[width=0.37\textwidth]{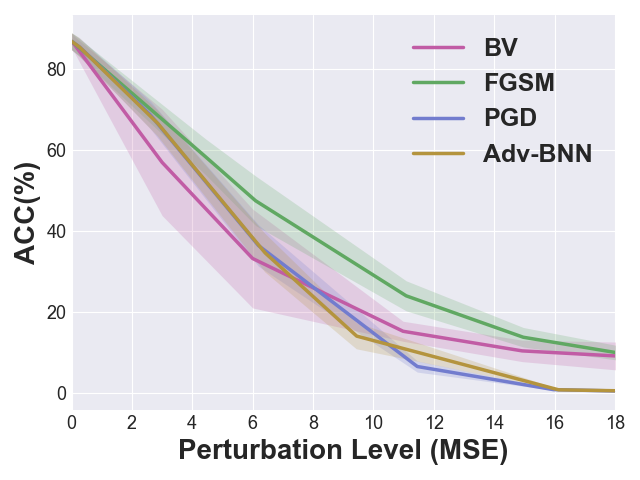}&
		\includegraphics[width=0.37\textwidth]{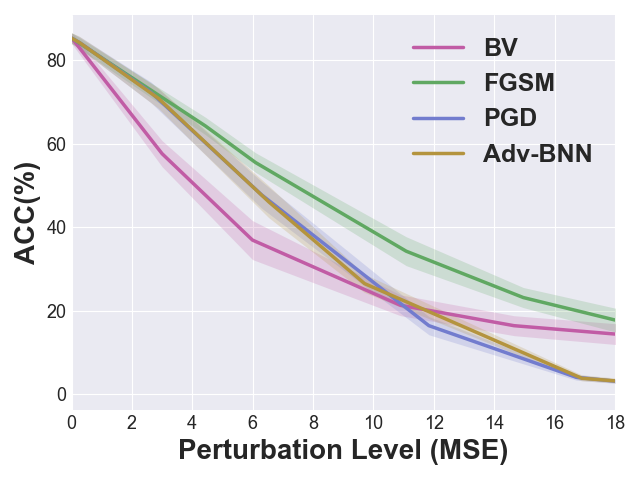}\\
		(a)  MobileNetV2--Basic Training & (b) MobileNetV2--FGSM Adv-Training&  (c) MobileNetV2--BV Adv-Training& (d) MobileNetV2--PGD Adv-Training

\end{tabular}
\caption{The effect of training with FGSM, PGD, and BV on bias and variance of the accuracy on CIFAR-10 dataset against adversarial attacks. Basic training refers to training without using adversarial examples.}
\label{fig:acc-cifar10}
\vspace{-0.5cm}
\end{figure*}

Figure~\ref{fig:acc-cifar10} also demonstrates the comparison results of the proposed BV adversarial attack compared to three state-of-the-art adversarial machine learning algorithms including FGSM~\cite{goodfellow2014explaining}, PGD~\cite{madry2017towards} and Adv-BNN~\cite{liu2018adv} methods. The methods are evaluated by different adversarial training approaches shown in columns (b)--(d). As seen, the proposed BV algorithm provides higher success rates in fooling the networks within lower perturbation levels.
It is worth to note that PGD and Adv-BNN are iterative algorithms in generating the perturbation while FGSM and the proposed BV algorithm generate the perturbation in one step. As such, PGD and Adv-BNN methods have a higher success rate when the perturbation level is increased.
Moreover, the computational complexity of the proposed  BV adversarial attacks is as low as FGSM's computational complexity, while it outperforms FGSM significantly. Furthermore, the proposed BV method is $k$ times faster than multi-step attacks such as PGD where $k$ is the number of inner iterations of PGD attack, while BV produces higher success rate in lower perturbation levels. This benefits the model even  more  when the proposed method is being used in the adversarial training step as it reduces the training time significantly.

The details of the experimental setup including the hyper-parameters setting and the results for CIFAR-100 dataset are reported in the supplementary material.

\vspace{-0.5cm}
\section{Conclusion}
\vspace{-0.25cm}
In this paper we studied  the effect of adversarial machine learning on a model's bias and variance. We proposed a new set of theorems which decompose the effect of adversarial perturbations on machine learning models trained with two well-known loss functions of MSE and cross-entropy. The new derivations showed when to expect the maximum increase in the bias and variance of the model facing adversarial machine leanings. While the theorems verify the previous findings in this field which the model is vulnerable in the opposite direction of gradient of loss function, the proposed theorems can quantify  what is the best direction for adversarial perturbation to maximize the effect. The proposed theorems can help us to better understand the effect of adversarial machine learning algorithms in the field of deep neural networks and hopefully benefit the field to resolve this concern.

Moreover, motivated by the new findings, we proposed a new adversarial machine learning method so-call BV adversarial attack which can fool deep neural network with  a reasonable success rate compared to other state-of-the-art algorithms with a less computational complexity.  Experimental results showed that the proposed attack can fool the network with a higher success rate while adds a lower perturbations to the input image compared to the state-of-the-art methods.
\section{ Broader Impact}
Adversarial machine learning poses serious challenges for deep neural networks~\cite{akhtar2018threat,goodfellow2014explaining} to behave erroneously in the presence of adversarial perturbation. With the rise in prevalence of deep neural networks being used in real-world mission-critical applications ranging from video surveillance and autonomous driving cars to biometric recognition and financial trading, adversarial machine learning can cause serious negative socioethical concerns as well as physical harm when leveraged with malicious intent.  In this work, we present a theory to help in understanding the impact of adversarial machine learning on both the variance and bias of the system, and for the first time illustrate how adversarial perturbations can manipulate the variance of the system besides its bias. We believe that these types of theoretical insights will give us a deeper understanding how these mechanisms lead deep neural networks to become vulnerable facing adversarial perturbation. Knowing how deep neural networks fail under adversarial machine learning will allow the community to discover new ways to defend against them and improve their robustness to such perturbations in order to build more reliable deep neural networks to use in real-world scenarios that impact society at large.
%

%
%
\section{Theorems and Derivations}

Theorems 2 and 3 in the main manuscript derive the bias-variance trade-off for the MSE and cross-entropy loss functions in the presence of an adversarial perturbation. Theorem 4 derives the bias-variance decomposition of the cross-entropy loss function of a prediction model constructed by the set of functions in the presence of adversarial  perturbation, which provides a direction which maximizes the loss. This new decomposition inspired  the proposed BV adversarial machine learning algorithm. Below the detailed derivation and proofs of the proposed theorems are provided.

\textbf{Proof of Theorem 2:}\\
Given the adversarial perturbation $\beta(x)$ with the condition $l_{\infty}\big(\beta(x)\big)< \delta$ and the assumption that $f(x+\beta(x))=f(x)$, the MSE loss can be decomposed as follows:

\begin{align}\label{eq3}
	\EX_{x,\mathcal{D},\gamma}\left[(y-\hat{f}(x+\beta(x)))^2\right]&=\EX_{x,\mathcal{D},\gamma}\left[(f(x)-\hat{f}(x+\beta(x))+\gamma)^2\right]\nonumber\\&=
	\EX_{x,\mathcal{D},\gamma}\left[(f(x)-\bar{f}(x+\beta(x))+\bar{f}(x+\beta(x))-\hat{f}(x+\beta(x))+\gamma)^2\right]\nonumber\\&=
	\EX_{x,\mathcal{D}}\left[(f(x)-\bar{f}(x+\beta(x)))^2\right]+\EX_{x,\mathcal{D}}\left[(\hat{f}(x+\beta(x))-\bar{f}(x+\beta(x)))^2	\right]+\sigma^2_\gamma	\nonumber
	\\&+2\EX_{\gamma}[\gamma]\EX_{x,\mathcal{D}}\left[(f(x)-\bar{f}(x+\beta(x))+(\bar{f}(x+\beta(x))-\hat{f}(x+\beta(x)))\right]\nonumber
	\\&+2\EX_{x}\left[(f(x)-\bar{f}(x+\beta(x))\times\EX_{\mathcal{D}}\left[(\bar{f}(x+\beta(x))-\hat{f}(x+\beta(x)))\right]\right]\nonumber
	\\&=\EX_{x,\mathcal{D}}\left[(f(x)-\bar{f}(x+\beta(x)))^2\right]+\EX_{x,\mathcal{D}}\left[(\hat{f}(x+\beta(x))-\bar{f}(x+\beta(x)))^2	\right]+\sigma^2_\gamma
	\nonumber
	\\&\approx \EX_{x,\mathcal{D}}\left[\left(f(x)-\bar{f}(x)-\nabla\bar{f}(x)^T\beta(x)\right)^2\right]+\EX_{x,\mathcal{D}}\left[(\hat{f}(x+\beta(x))-\bar{f}(x+\beta(x)))^2	\right]+\sigma^2_\gamma
\end{align}
In \eqref{eq3}, we are using the fact that $\EX_{\gamma}[\gamma]=0$ and $\EX_{\mathcal{D}}\left[(\bar{f}(x+\beta(x))-\hat{f}(x+\beta(x)))\right]=0$.

Also for the term $\EX_{x,\mathcal{D}}\left[(\hat{f}(x+\beta(x))-\bar{f}(x+\beta(x)))^2	\right]$ in \eqref{eq3}, by using Taylor polynomial of order one \cite{thomas2010thomas}, we have:
\begin{small}
\begin{align}\label{var1}
    \EX_{x,D}&\left[(\hat{f}(x+\beta(x))-\bar{f}(x+\beta(x)))^2]
    \right]\approx\EX_{x,D}\left[\left(\hat{f}(x)-\bar{f}(x)\right)^2 +2(\hat{f}(x)-\bar{f}(x))(\nabla\hat{f}(x)-\nabla\bar{f}(x))^T\beta(x) \right]\nonumber\\
    &=Var[\hat{f}]+\EX_{x,D}\left[2(\hat{f}(x)-\bar{f}(x))(\nabla\hat{f}(x)-\nabla\bar{f}(x))^T\beta(x)\right]
\end{align}
	\end{small}
Putting together \eqref{var1} and \eqref{eq3}, we have:
\begin{small}
 \begin{align} \label{eq7}
 &    \EX_{x,\mathcal{D},\gamma}\left[(y-\hat{f}(x+\beta(x)))^2\right]\approx\EX_{x,\mathcal{D}}[(f(x)-\bar{f}(x)-c_x)^2]+Var[\gamma]+Var[\hat{f}]+\EX_{x,\mathcal{D}}[c_x'] \\
&\text{where,   } \;\;\;c_x =\nabla\bar{f}(x)^T\beta(x)  \;\;\text{and,} \;\; c_x'= 2\Big(\hat{f}(x)-\bar{f}(x)\Big) \Big(\Big(\nabla\hat{f}(x)-\nabla\bar{f}(x)\Big)^T\beta(x)\Big)
\label{eq:cx_tem}
\end{align}
	\end{small}
\hfill$\blacksquare$\\

\textbf{Proof of  Theorem 3:}\\
Assuming that for input $x$, the ground truth class is $t_x$. By using Taylor polynomial of order one \cite{thomas2010thomas}, the loss can be decomposed as follow:

\begin{align}\label{cross33}
    L(\pi,\hat{\pi})&=\EX\limits_{x,\mathcal{D}}\left[D_{KL}(\pi(x)||{\pi^*}(x+\beta(x)))+D_{KL}({\pi}^*(x+\beta(x))||\hat{\pi}(x+\beta(x)))\right]\nonumber\\
    &=-\EX\limits_{x}\left[\log{\pi}^*_{t_x}(x+\beta(x))\right]-\EX\limits_{x,\mathcal{D}}\left[\sum_{i=1}^c{\pi}^*_i(x+\beta(x))\log\frac{\hat{\pi}_i(x+\beta(x))}{{\pi}^*_i(x+\beta(x))}\right]\nonumber\\
    &=-\EX\limits_{x}\left[\log{\pi}^*_{t_x}(x)\right]-\EX\limits_{x,D}\left[\sum_{i=1}^c{\pi}^*_i(x)\log\frac{\hat{\pi}_i(x)}{{\pi}^*_i(x)}\right]\nonumber\\
    &-\EX\limits_{x}\left[\big{(}\nabla_x\log{\pi}^*_{t_x}(x)\big{)}^T\beta(x)\right]-\EX\limits_{x,\mathcal{D}}\left[\sum_{i=1}^c\big{(}\nabla_x{\pi}^*_i(x)\log\frac{\hat{\pi}_i(x)}{{\pi}^*_i(x)}\big{)}^T\beta(x)\right]\nonumber\\
    &=\EX\limits_{x,\mathcal{\mathcal{D}}}\big[D_{KL}(\pi(x)||{\pi}^*(x))+D_{KL}({\pi}^*(x)||\hat{\pi}(x))\big]\nonumber\\&-\EX\limits_{x}\left[\big{(}\nabla_x\log{\pi}^*_{t_x}(x)\big{)}^T\beta(x)\right]
    -\EX\limits_{x,\mathcal{D}}\left[\sum_{i=1}^c\big{(}\nabla_x{\pi}^*_i(x)\log\frac{\hat{\pi}_i(x)}{{\pi}^*_i(x)}\big{)}^T\beta(x)\right]
\end{align}
\hfill$\blacksquare$\\
\textbf{Proof of Theorem 4:}\\
Here, as we have the assumption that the perturbation $\beta(x)$ does not change the ground-truth class of $x$, by using Taylor polynomial of order one \cite{thomas2010thomas}, the loss can be decomposed as follows:
\begin{small}
\begin{align}\label{cross1}
    &\EX_{x,\mathcal{D}}\left[\sum_{i=1}^c\left(-\log(\frac{\hat{f}_i(x+\beta(x))}{\hat{f}_1(x+\beta(x))+...+\hat{f}_c(x+\beta(x))})\right)\mathbbmss{1}_{f(x)=i}\right]\nonumber\\ &=\sum_{i=1}^c\EX_{x\in X_i,\mathcal{D}}\left[-\log\big(\hat{f}_i(x+\beta(x))\big)+\log\big(\hat{f}_1(x+\beta(x))+...+\hat{f}_c(x+\beta(x))\big)\right]\nonumber\\&\approx
    \sum_{i=1}^c\EX_{x \in X_i,\mathcal{D}}\left[-\log(\hat{f}_i(x))+\log(\hat{f}_1(x)+...+\hat{f}_c(x))\right]
    +
    \sum_{i=1}^c\EX_{x \in X_i,\mathcal{D}}\left[\left(-\frac{\nabla\hat{f}_i(x)}{\hat{f}_i(x)}+\frac{\nabla\hat{f}_1(x)+...+\nabla\hat{f}_c(x)}{\hat{f}_1(x)+...+\hat{f}_c(x)}\right)^T\beta(x)\right]
    \nonumber\\&=
    \sum_{i=1}^c\EX_{x \in X_i,\mathcal{D}}\left[-\log(\hat{f}_i(x))+\log(\hat{f}_1(x)+...+\hat{f}_c(x))\right]
    +\sum_{i=1}^c\EX_{x\in X_i,\mathcal{D}}\left[-{\nabla} log\left(\frac{\hat{f}_i(x)}{\hat{f}_1(x)+...+\hat{f}_c(x)}\right)^T\beta(x)\right]
\end{align}
	\end{small}
\hfill$\blacksquare$\\
\section{Experimental Setup}

The image samples of both CIFAR-10 and CIFAR-100 datasets ~\cite{krizhevsky2009learning}, are normalized by the mean $[0.4914, 0.4822, 0.4465]$ and standard deviation $[0.2023, 0.1994, 0.2010]$ for the three RGB channels.
The experimental results were done for all networks including ResNet18, ResNet34, ResNet50 \cite{he2016deep} and MobileNetV2 ~\cite{sandler2018mobilenetv2} in two different setups of  with  adversarial training and no adversarial training. The $\epsilon = 0.05$ and $\epsilon = 0.02$ were used for adversarial training with FGSM~\cite{goodfellow2014explaining} and PGD~\cite{madry2017towards}, respectively, while the number of iterations for generative PGD perturbation was set to 5. For adversarial training with BV, $\epsilon$ was set to 2.5. These epsilons are chosen based on the grid-search.
It is also worth to mention that to make a fair comparison of different adversarial training, it was made sure that adversarial training with different algorithms resulted to a similar amount $l_2$ norm perturbation on the input image (equal to the square root of MSE difference of the ground truth image and the perturbed one.).

For evaluating different method in the test time, and comparing the effect of adversarial perturbation on the trained models, the range of $\epsilon$ for each perturbation is determined such that  it injects the same  amount of perturbation (measured by the MSE difference of the ground truth image and the perturbed one) compared to other perturbations. Moreover, since adversarial trained models had more resilient against adversarial perturbation, we used higher range of $\epsilon$ to perform adversarial perturbation on the models in the test time compared to the model with no adversarial training. All experiments were repeated 5 times with 5 difference seed values to obtain reliable variances.

\section{Experimental Results}
\begin{figure*}
\vspace{-0.35cm}
\hspace*{-1.5in}
\footnotesize
\setlength{\tabcolsep}{0.01cm}
\centering
        \begin{tabular}{cccc}
        \includegraphics[width=0.37\textwidth]{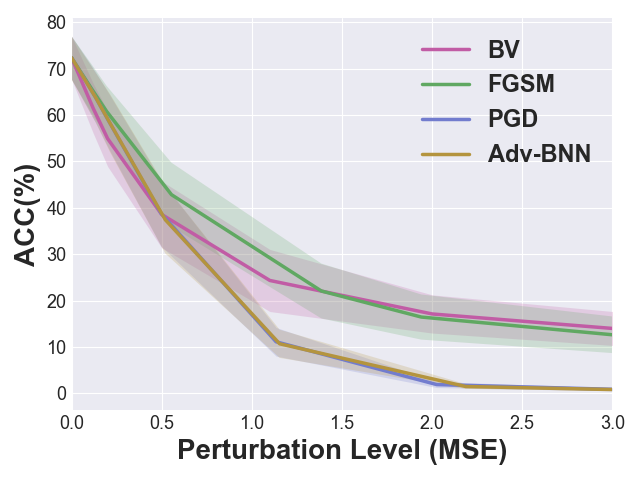}&
		\includegraphics[width=0.37\textwidth]{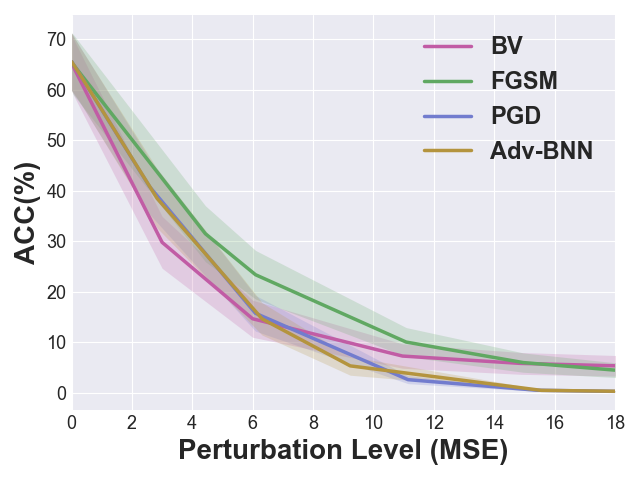}&
		\includegraphics[width=0.37\textwidth]{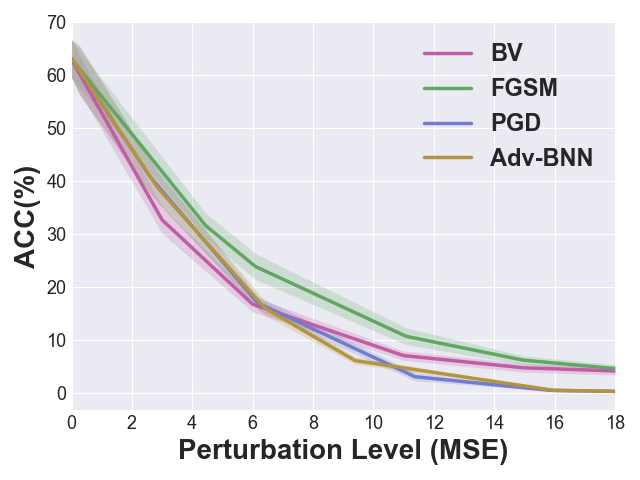}&
		\includegraphics[width=0.37\textwidth]{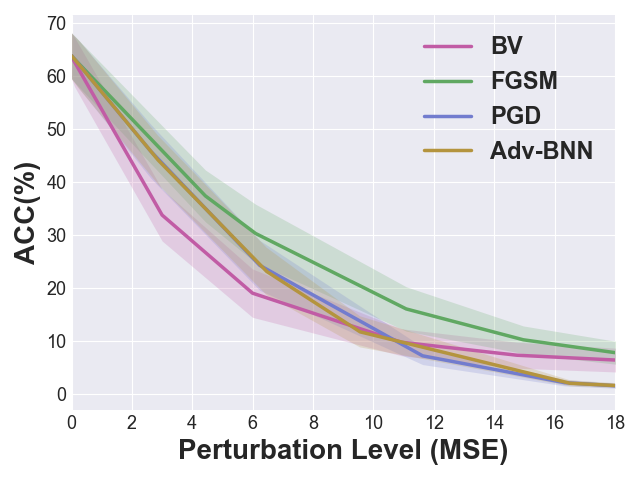}\\
		(a)  ResNet18--No Adversarial Training & (b) ResNet18--FGSM Adv-Training&  (c) ResNet18--BV Adv-Training& (d) ResNet18--PGD Adv-Training \\
		\includegraphics[width=0.37\textwidth]{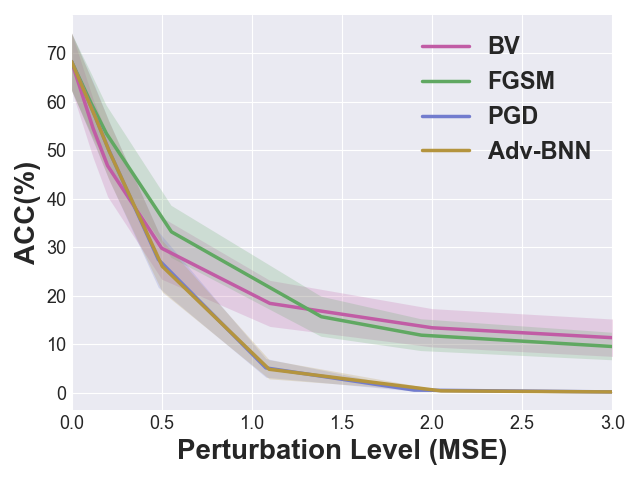}&
		\includegraphics[width=0.37\textwidth]{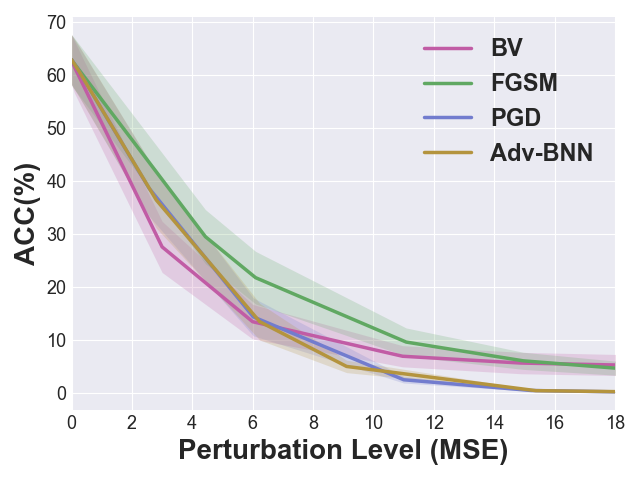}&
		\includegraphics[width=0.37\textwidth]{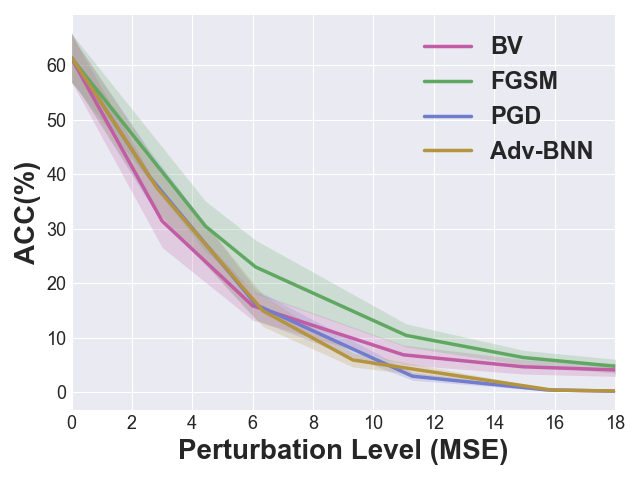}&
		\includegraphics[width=0.37\textwidth]{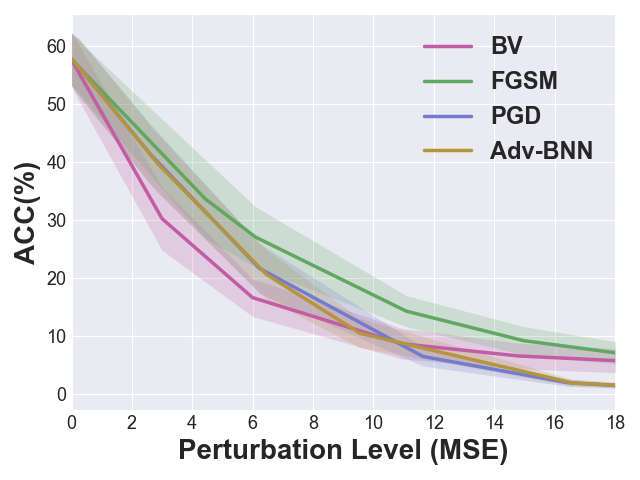}\\
		(a)  MobileNetV2--No Adversarial Training & (b) MobileNetV2--FGSM Adv-Training&  (c) MobileNetV2--BV Adv-Training& (d) MobileNetV2--PGD Adv-Training  \\
		\includegraphics[width=0.37\textwidth]{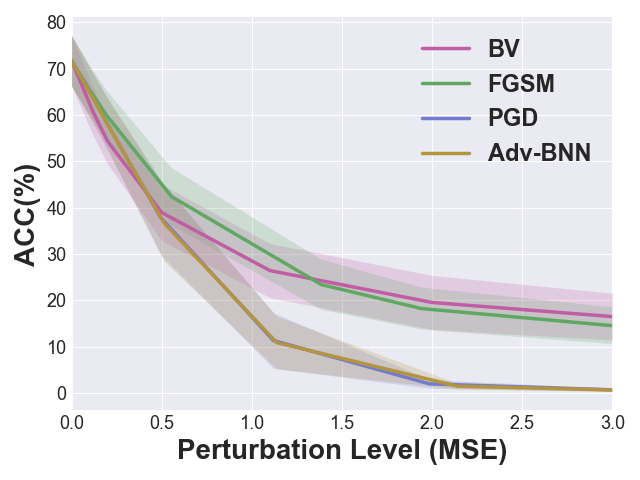}&
		\includegraphics[width=0.37\textwidth]{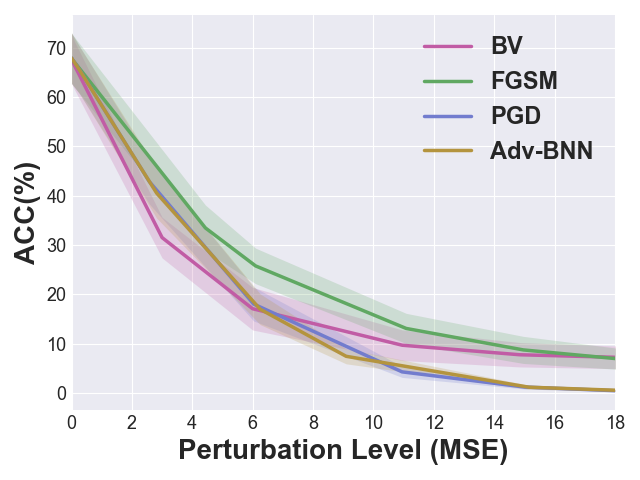}&
		\includegraphics[width=0.37\textwidth]{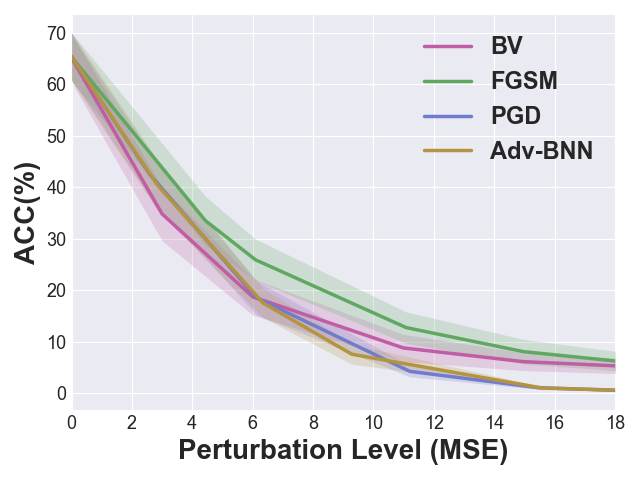}&
		\includegraphics[width=0.37\textwidth]{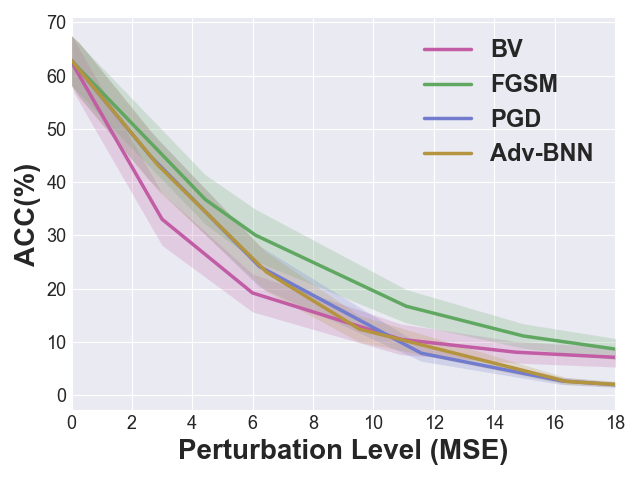}\\
		(a)  ResNet34--No Adversarial Training & (b) ResNet34--FGSM Adv-Training&  (c) ResNet34--BV Adv-Training& (d) ResNet34--PGD Adv-Training \\
		\includegraphics[width=0.37\textwidth]{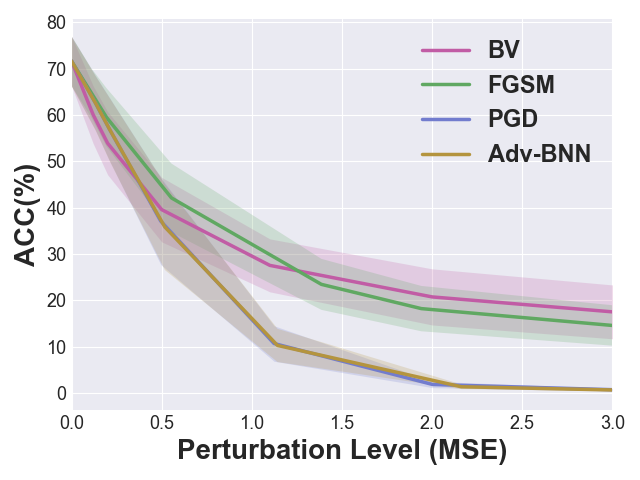}&
		\includegraphics[width=0.37\textwidth]{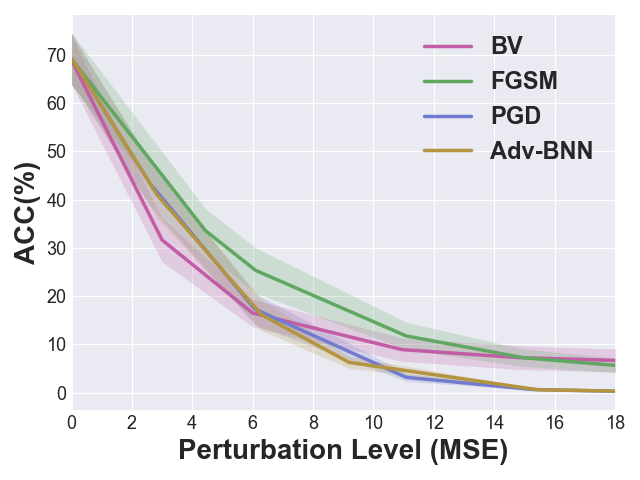}&
		\includegraphics[width=0.37\textwidth]{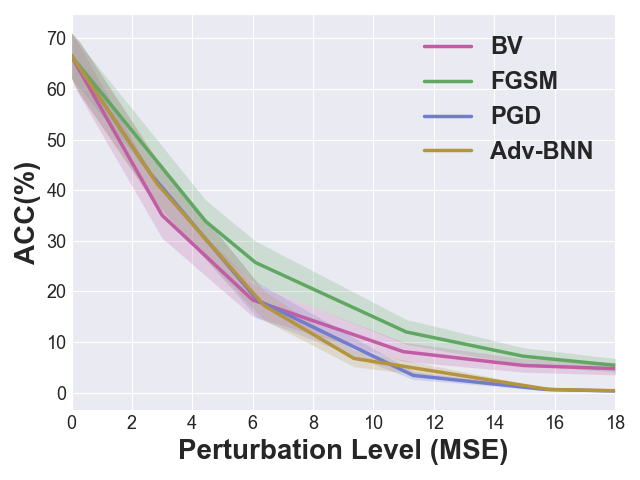}&
		\includegraphics[width=0.37\textwidth]{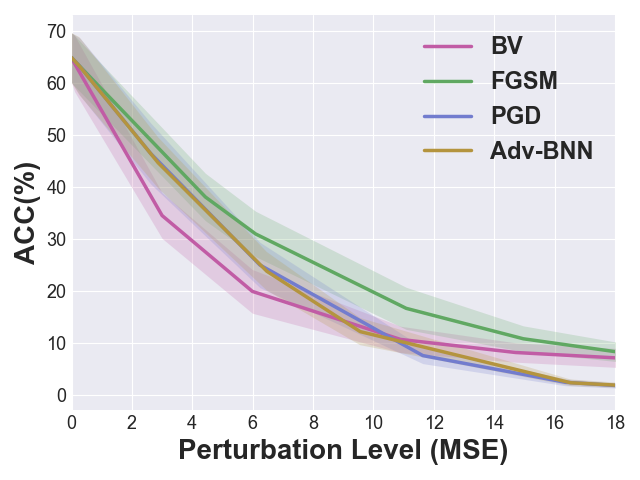}\\
		(a)  ResNet50--No Adversarial Training & (b) ResNet50--FGSM Adv-Training&  (c) ResNet50--BV Adv-Training& (d) ResNet50--PGD Adv-Training
\end{tabular}
\caption{The effect of training with FGSM, PGD, and BV on bias and variance of the accuracy on CIFAR-100 dataset against adversarial perturbations. }
\label{fig:acc-cifar100}
\vspace{-0.5cm}
\end{figure*}

\begin{figure*}
\vspace{-0.35cm}
\hspace*{-1.5in}
\footnotesize
\setlength{\tabcolsep}{0.01cm}
\centering
        \begin{tabular}{cccc}
        \includegraphics[width=0.37\textwidth]{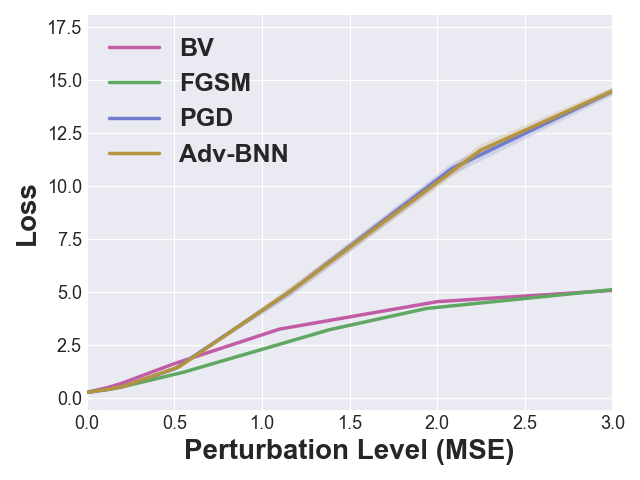}&
		\includegraphics[width=0.37\textwidth]{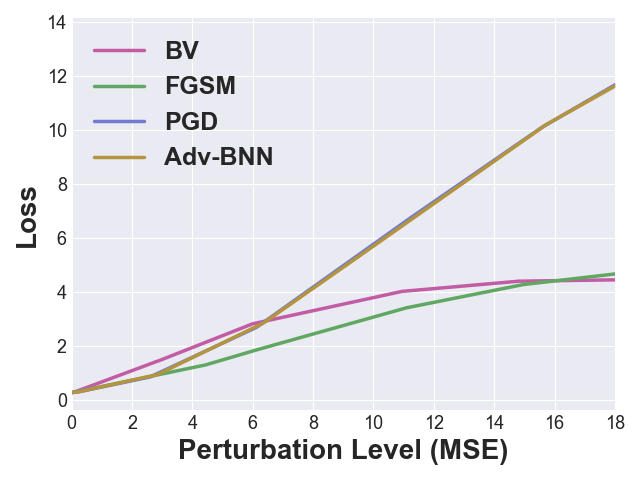}&
		\includegraphics[width=0.37\textwidth]{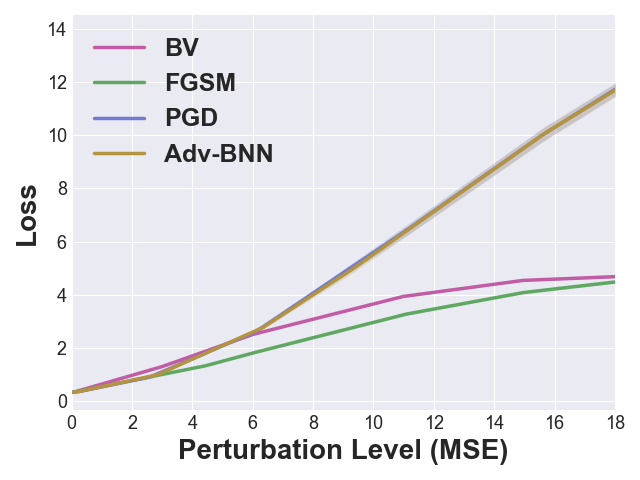}&
		\includegraphics[width=0.37\textwidth]{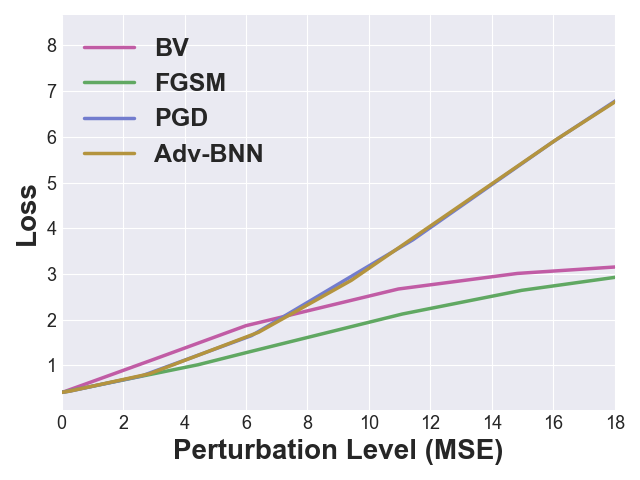}\\
		(a)  ResNet18--No Adversarial Training & (b) ResNet18--FGSM Adv-Training&  (c) ResNet18--BV Adv-Training& (d) ResNet18--PGD Adv-Training \\
		\includegraphics[width=0.37\textwidth]{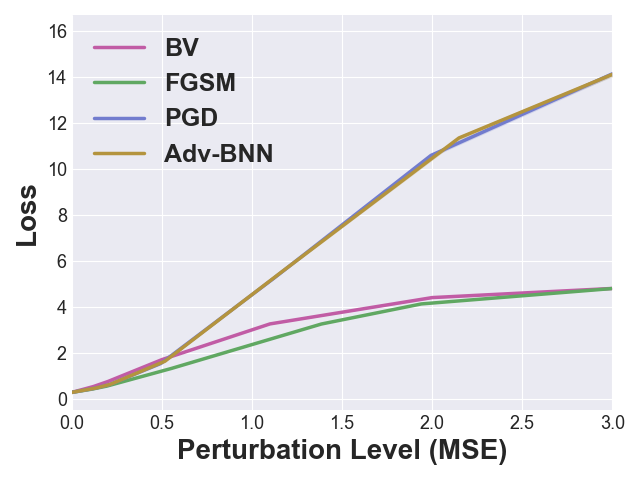}&
		\includegraphics[width=0.37\textwidth]{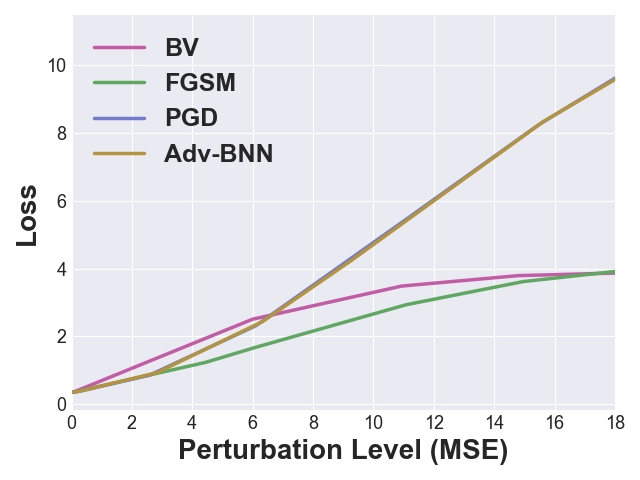}&
		\includegraphics[width=0.37\textwidth]{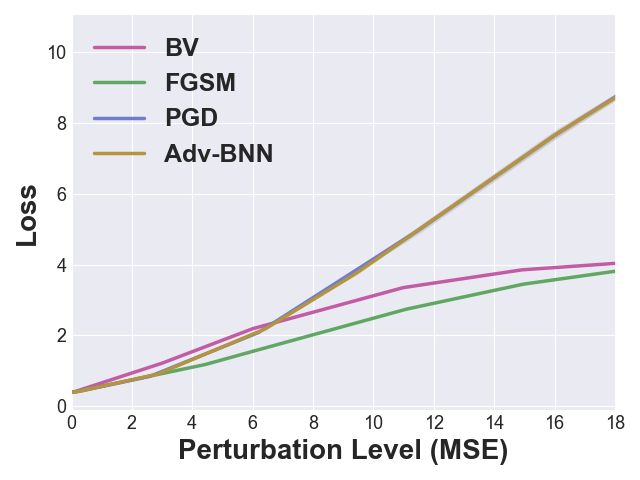}&
		\includegraphics[width=0.37\textwidth]{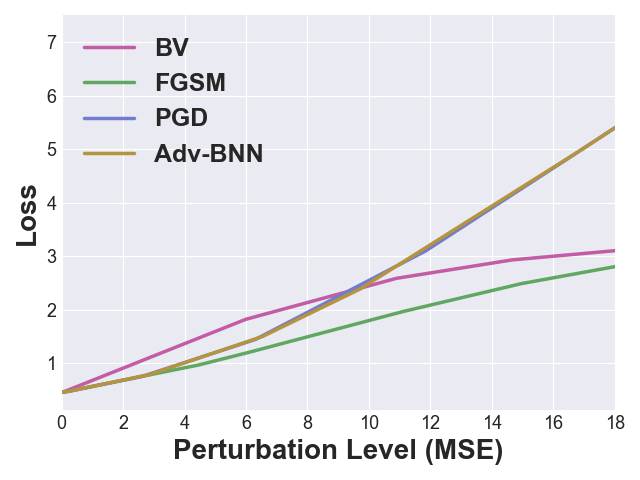}\\
		(a)  MobileNetV2--No Adversarial Training & (b) MobileNetV2--FGSM Adv-Training&  (c) MobileNetV2--BV Adv-Training& (d) MobileNetV2--PGD Adv-Training  \\
		\includegraphics[width=0.37\textwidth]{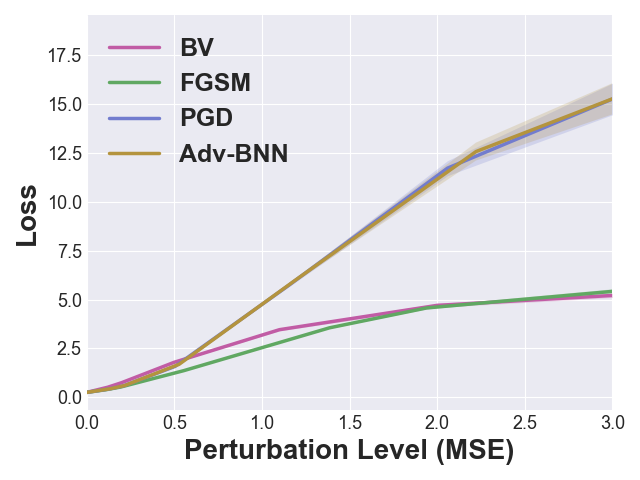}&
		\includegraphics[width=0.37\textwidth]{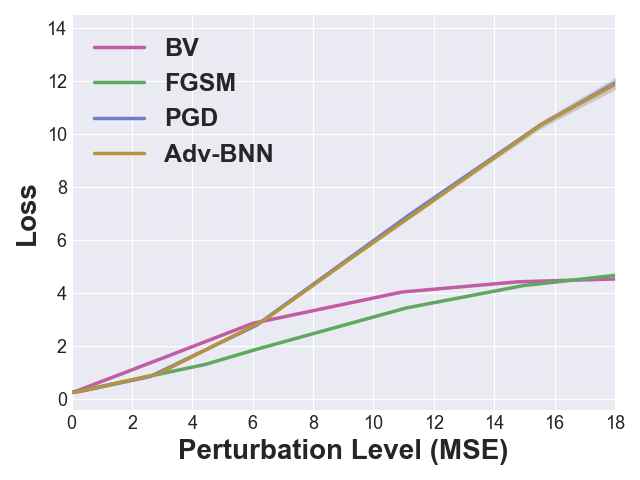}&
		\includegraphics[width=0.37\textwidth]{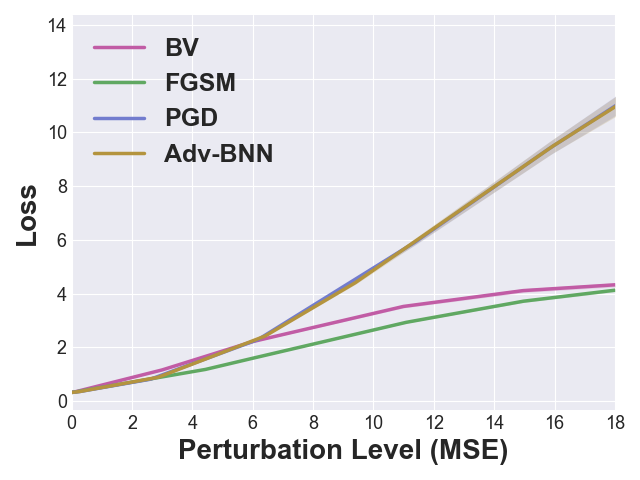}&
		\includegraphics[width=0.37\textwidth]{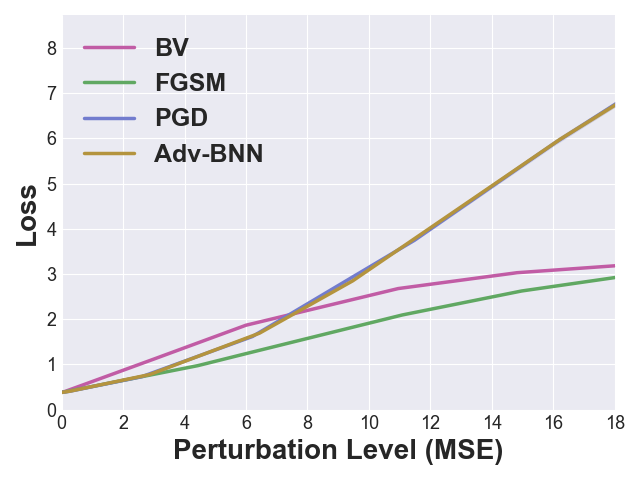}\\
		(a)  ResNet34--No Adversarial Training & (b) ResNet34--FGSM Adv-Training&  (c) ResNet34--BV Adv-Training& (d) ResNet34--PGD Adv-Training \\
		\includegraphics[width=0.37\textwidth]{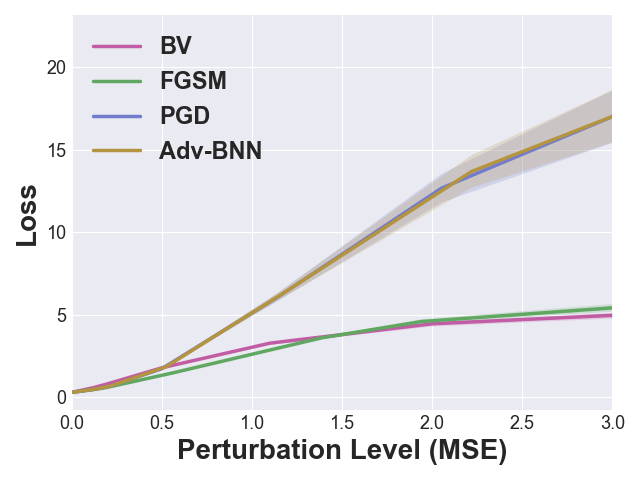}&
		\includegraphics[width=0.37\textwidth]{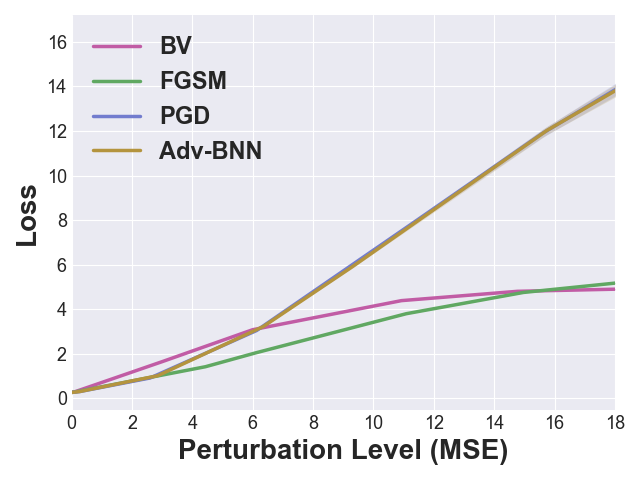}&
		\includegraphics[width=0.37\textwidth]{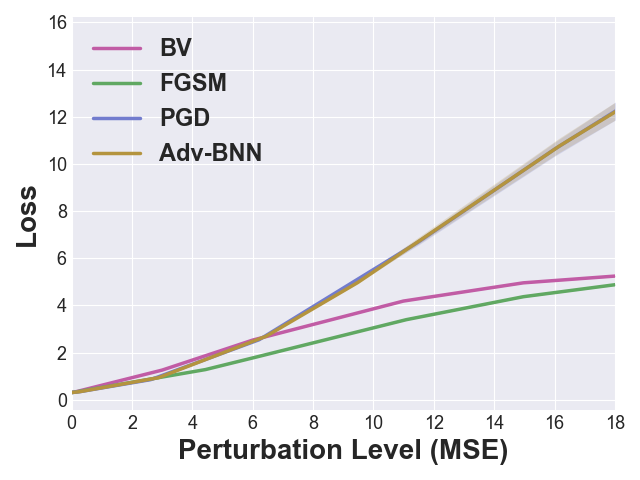}&
		\includegraphics[width=0.37\textwidth]{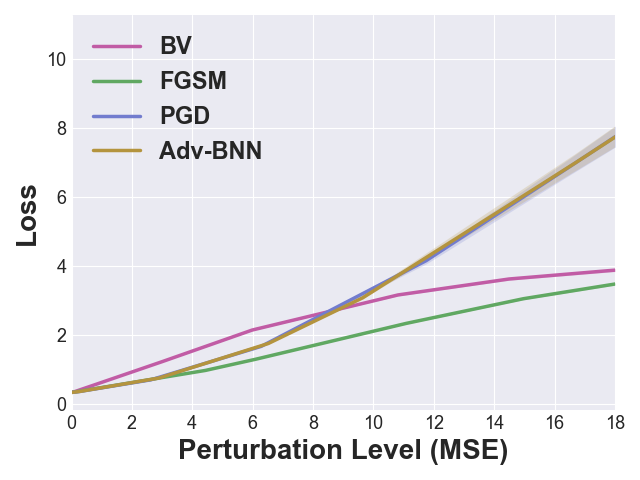}\\
		(a)  ResNet50--No Adversarial Training & (b) ResNet50--FGSM Adv-Training&  (c) ResNet50--BV Adv-Training& (d) ResNet50--PGD Adv-Training
\end{tabular}
\caption{The effect of training with FGSM, PGD, and BV on bias and variance of the Loss on CIFAR-10 dataset against adversarial perturbations. }
\label{fig:loss-cifar10}
\vspace{-0.5cm}
\end{figure*}

\begin{figure*}
\vspace{-0.35cm}
\hspace*{-1.5in}
\footnotesize
\setlength{\tabcolsep}{0.01cm}
\centering
        \begin{tabular}{cccc}
        \includegraphics[width=0.37\textwidth]{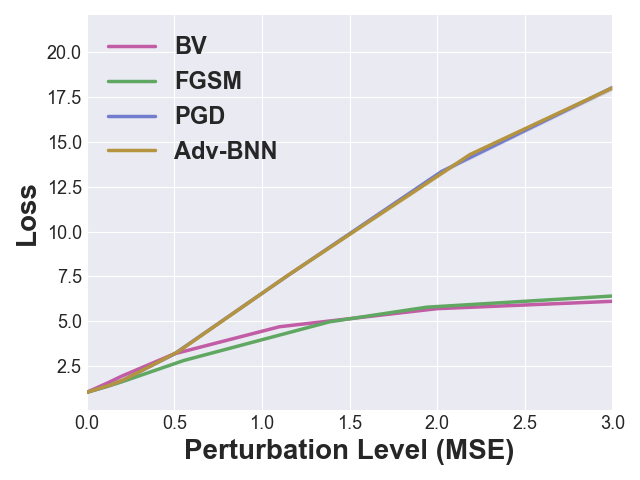}&
		\includegraphics[width=0.37\textwidth]{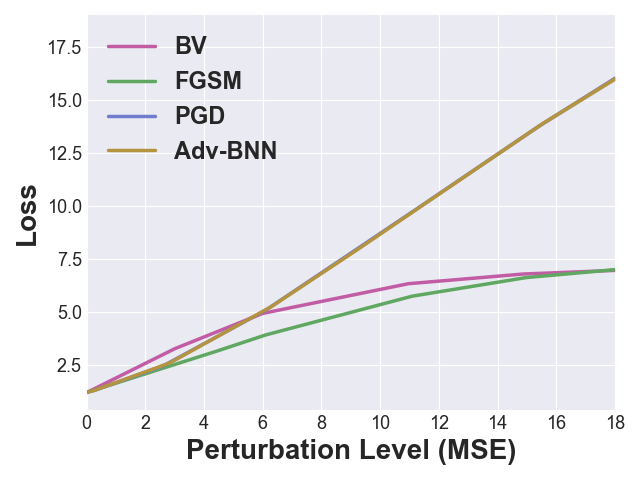}&
		\includegraphics[width=0.37\textwidth]{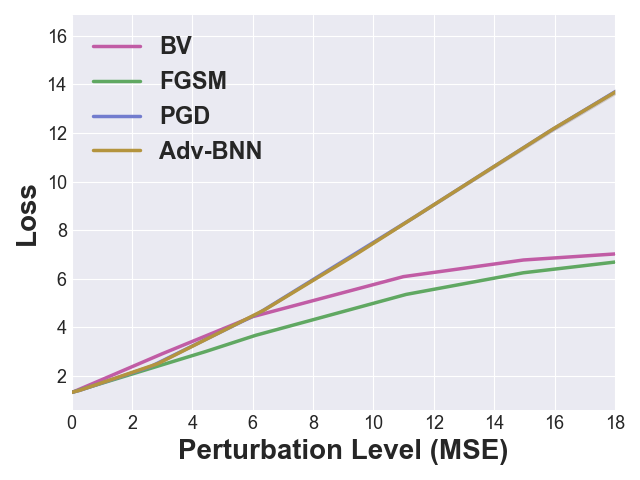}&
		\includegraphics[width=0.37\textwidth]{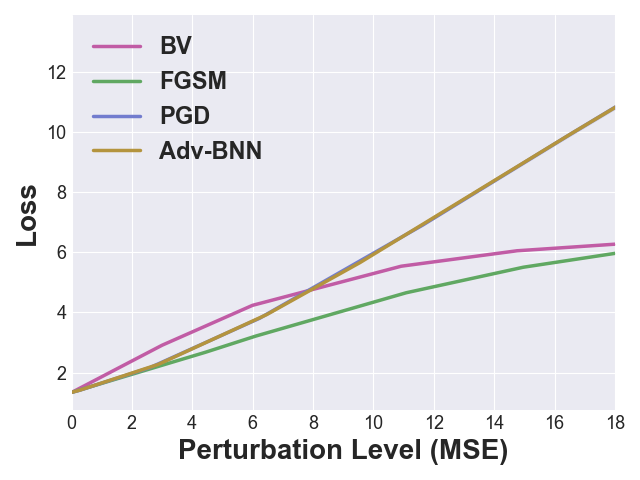}\\
		(a)  ResNet18--No Adversarial Training & (b) ResNet18--FGSM Adv-Training&  (c) ResNet18--BV Adv-Training& (d) ResNet18--PGD Adv-Training \\
		\includegraphics[width=0.37\textwidth]{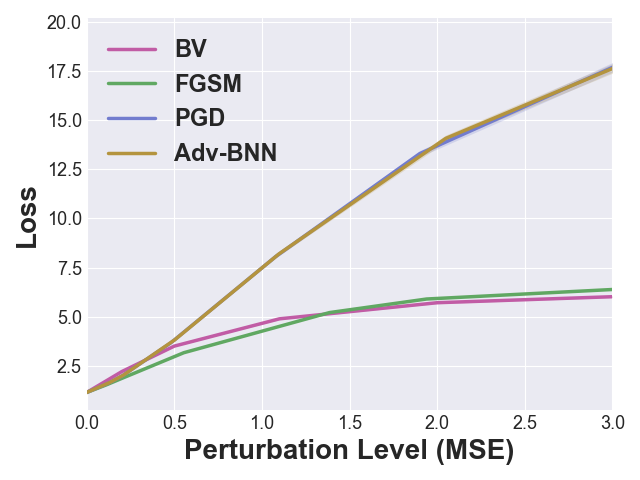}&
		\includegraphics[width=0.37\textwidth]{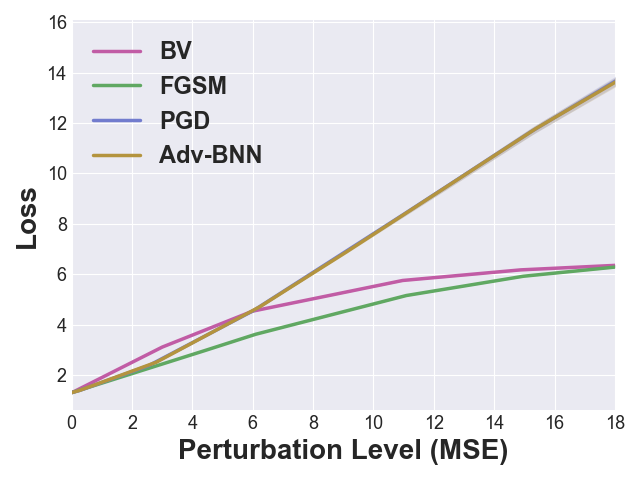}&
		\includegraphics[width=0.37\textwidth]{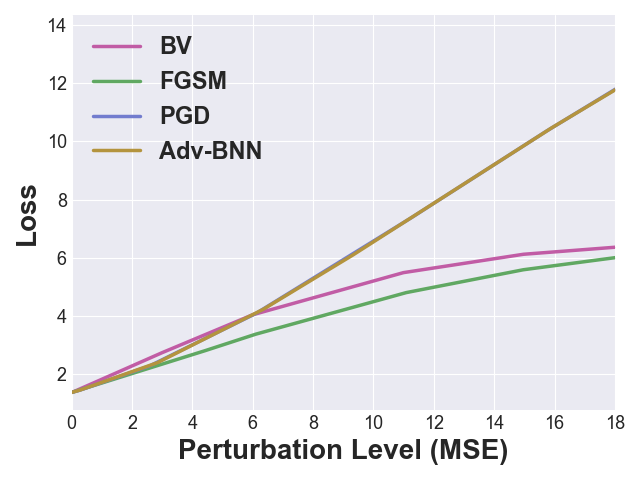}&
		\includegraphics[width=0.37\textwidth]{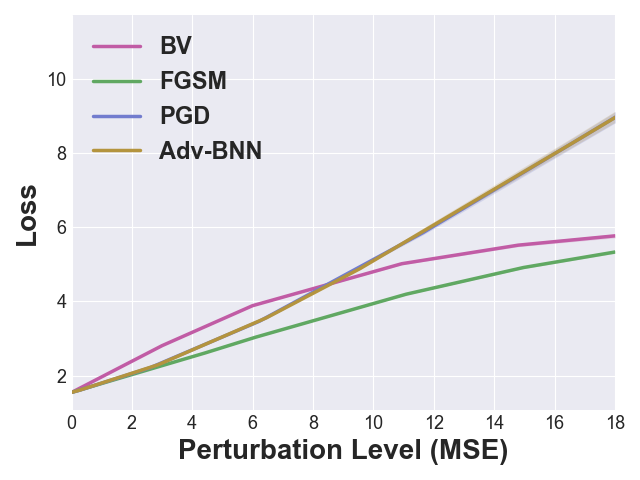}\\
		(a)  MobileNetV2--No Adversarial Training & (b) MobileNetV2--FGSM Adv-Training&  (c) MobileNetV2--BV Adv-Training& (d) MobileNetV2--PGD Adv-Training  \\
		\includegraphics[width=0.37\textwidth]{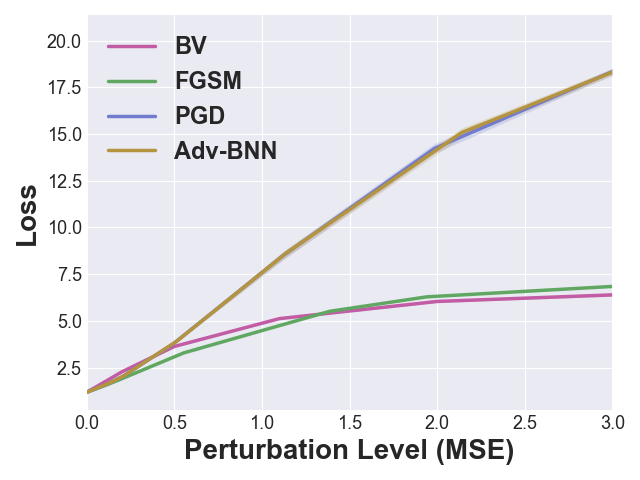}&
		\includegraphics[width=0.37\textwidth]{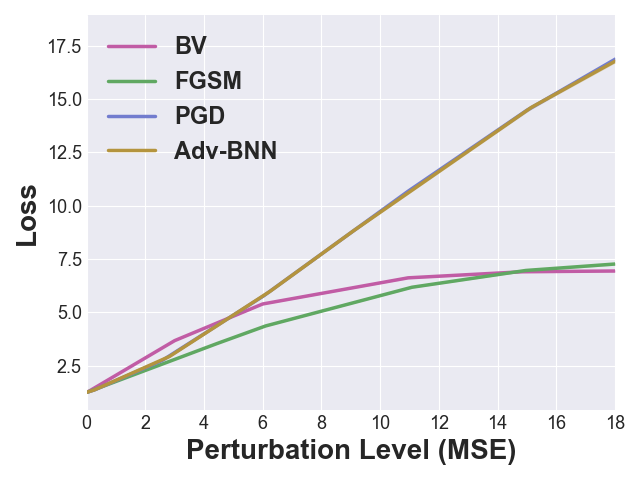}&
		\includegraphics[width=0.37\textwidth]{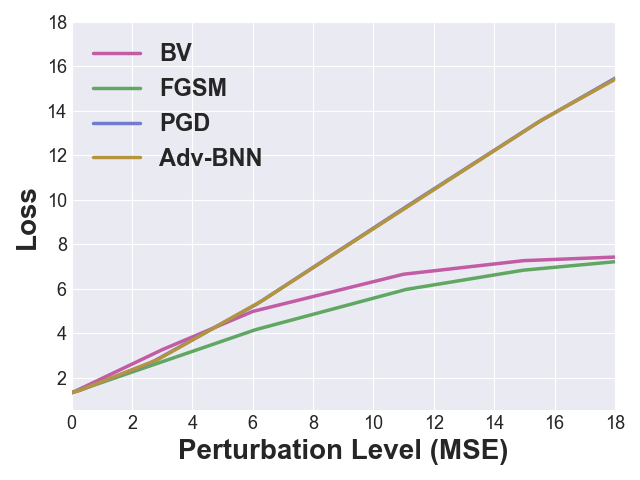}&
		\includegraphics[width=0.37\textwidth]{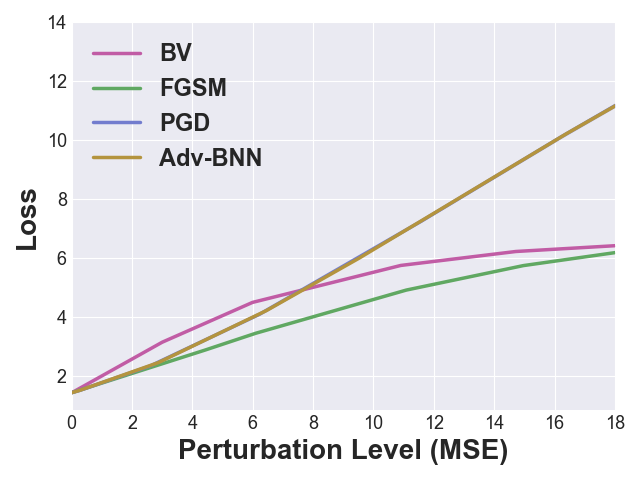}\\
		(a)  ResNet34--No Adversarial Training & (b) ResNet34--FGSM Adv-Training&  (c) ResNet34--BV Adv-Training& (d) ResNet34--PGD Adv-Training \\
		\includegraphics[width=0.37\textwidth]{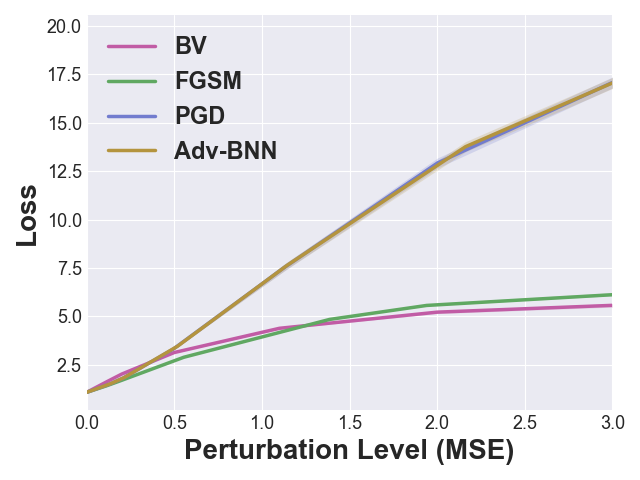}&
		\includegraphics[width=0.37\textwidth]{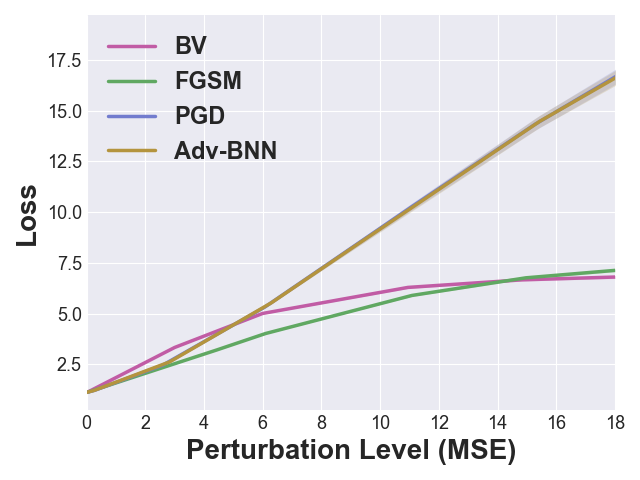}&
		\includegraphics[width=0.37\textwidth]{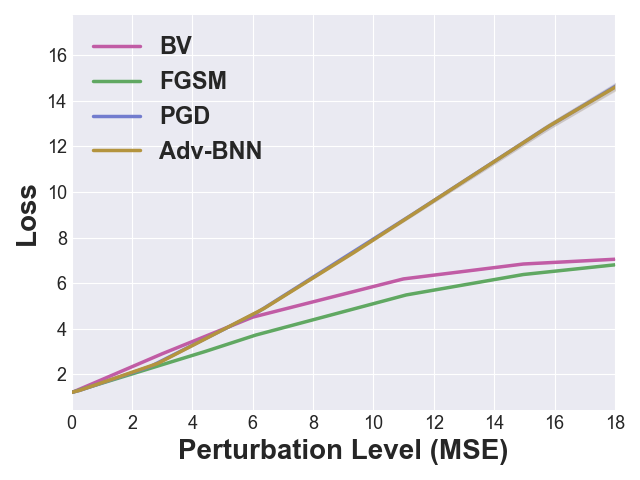}&
		\includegraphics[width=0.37\textwidth]{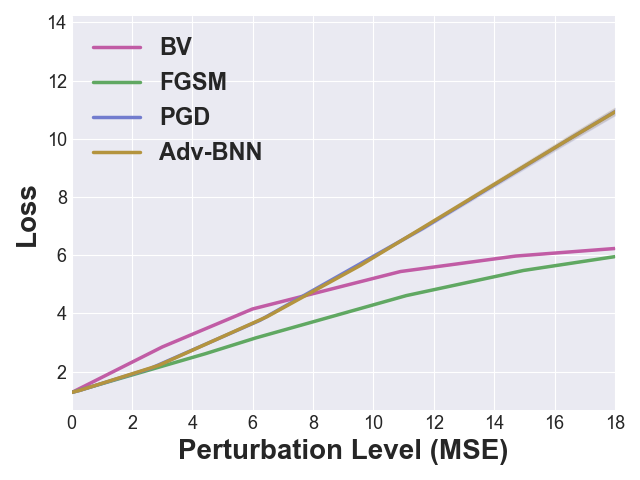}\\
		(a)  ResNet50--No Adversarial Training & (b) ResNet50--FGSM Adv-Training&  (c) ResNet50--BV Adv-Training& (d) ResNet50--PGD Adv-Training
\end{tabular}
\caption{The effect of training with FGSM, PGD, and BV on bias and variance of the Loss on CIFAR-100 dataset against adversarial perturbations. }
\label{fig:loss-cifar100}
\vspace{-0.5cm}
\end{figure*}

Figures~\ref{fig:acc-cifar10},~\ref{fig:acc-cifar100},~\ref{fig:loss-cifar10} and~\ref{fig:loss-cifar100}, shows the comprehensive experimental results for two datasets of CIFAR-10 and CIFAR-100 against 4 different deep neural network models.  To better analyze the impact of adversarial perturbation on the model's bias and variance, the changes in both accuracy and loss values for the models are reported given different perturbation levels. Figures~\ref{fig:acc-cifar10} and~\ref{fig:acc-cifar100} depict the impact of adversarial perturbation on the accuracy of the models for CIFAR-10 and CIFAR-100 while the next two Figures reports the changes in the loss function for the same experiments.

Column (a) in all Figures shows the performance of models with no adversarial training while the other Columns demonstrates the robustness of models against different adversarial machine learning algorithms when they are trained with   FGSM (Column (b)),  BV (Column (c)) and PGD adversarial training (Column (d)).  As discussed in the main paper,  the adversarial training makes the model more resilient against adversarial perturbation and the accuracy drops  slower compared to when there is no adversarial training. It is also evident by the results that the variance of the accuracy of different models (Figures~\ref{fig:acc-cifar10} and~\ref{fig:acc-cifar100}), first starts to increase and then decreases since the perturbation with very high perturbation causes almost zero accuracy which results in a small variance. On the other hand, for loss  values (Figures~\ref{fig:loss-cifar10} and~\ref{fig:loss-cifar100}), the variance in almost all cases keep increasing. The reason why the changes in the variances cannot be  clearly seen in the plots for the loss values is that it is small compared to the increase in the bias of the loss.

The experimental results show that the BV algorithm is able to attack the targeted model and fool it properly which results to a higher drop in accuracy than other adversarial machine learning algorithms for smaller perturbations. Nevertheless, multi-step perturbations like PGD or Adv-BNN are more effective on higher perturbation levels. The experimental results demonstrates the effectiveness of the proposed adversarial perturbation algorithm while they also consistent with the proposed theorems in analyzing the deep neural networks' bias and variance against adversarial machine learning algorithms.

It is also wroth to mention that the experimental results are aligned with the findings introduced in~\cite{madry2017towards}. As seen, a network with higher capacity (e.g., ResNet-50) shows a higher level of robustness against multi-step attacks like PGD  and provide slower accuracy drop rate when the perturbation level is being increased. This is evident by Figures~\ref{fig:acc-cifar10} and~\ref{fig:acc-cifar100}.

\begin{figure*}
\vspace{-0.35cm}
\hspace*{-1.5in}
\footnotesize
\setlength{\tabcolsep}{0.01cm}
\centering
        \begin{tabular}{cccc}
        \includegraphics[width=0.37\textwidth]{images/CIFAR10/acc_base_18.png}&
		\includegraphics[width=0.37\textwidth]{images/CIFAR10/acc_fgsm_18.png}&
		\includegraphics[width=0.37\textwidth]{images/CIFAR10/acc_bv_18.png}&
		\includegraphics[width=0.37\textwidth]{images/CIFAR10/acc_pgd_18.png}\\
		(a)  ResNet18--No Adversarial Training & (b) ResNet18--FGSM Adv-Training&  (c) ResNet18--BV Adv-Training& (d) ResNet18--PGD Adv-Training \\
		\includegraphics[width=0.37\textwidth]{images/CIFAR10/acc_base_mobile.png}&
		\includegraphics[width=0.37\textwidth]{images/CIFAR10/acc_fgsm_mobile.png}&
		\includegraphics[width=0.37\textwidth]{images/CIFAR10/acc_bv_mobile.png}&
		\includegraphics[width=0.37\textwidth]{images/CIFAR10/acc_pgd_mobile.png}\\
		(a)  MobileNetV2--No Adversarial Training & (b) MobileNetV2--FGSM Adv-Training&  (c) MobileNetV2--BV Adv-Training& (d) MobileNetV2--PGD Adv-Training  \\
		\includegraphics[width=0.37\textwidth]{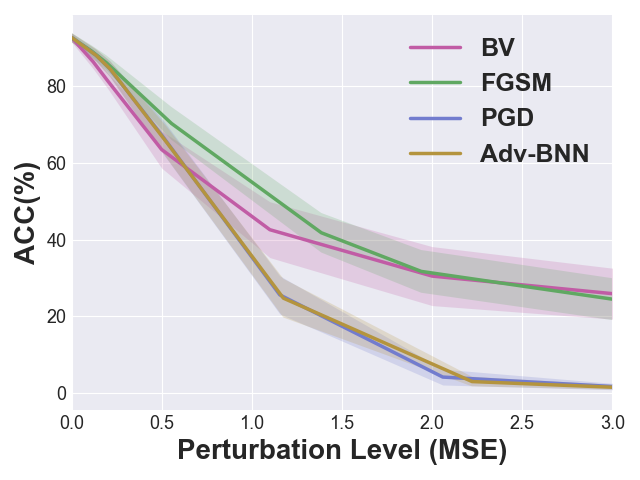}&
		\includegraphics[width=0.37\textwidth]{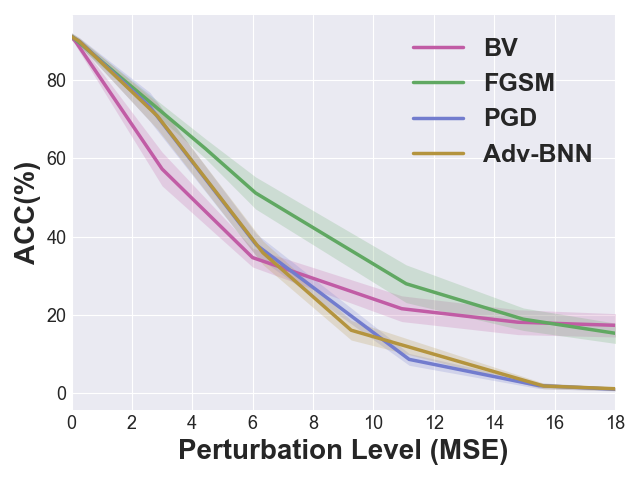}&
		\includegraphics[width=0.37\textwidth]{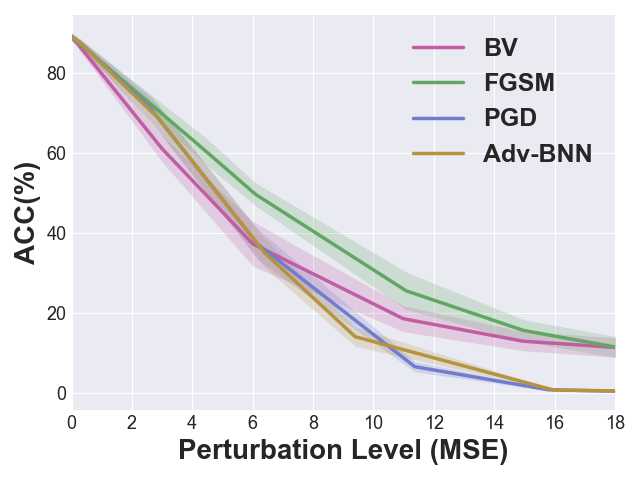}&
		\includegraphics[width=0.37\textwidth]{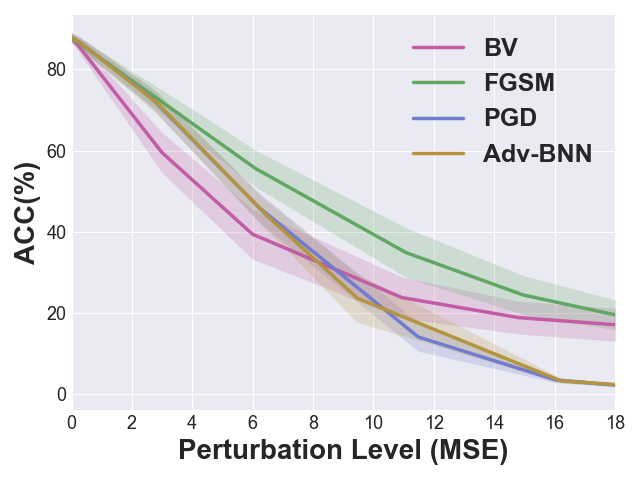}\\
		(a)  ResNet34--No Adversarial Training & (b) ResNet34--FGSM Adv-Training&  (c) ResNet34--BV Adv-Training& (d) ResNet34--PGD Adv-Training \\
		\includegraphics[width=0.37\textwidth]{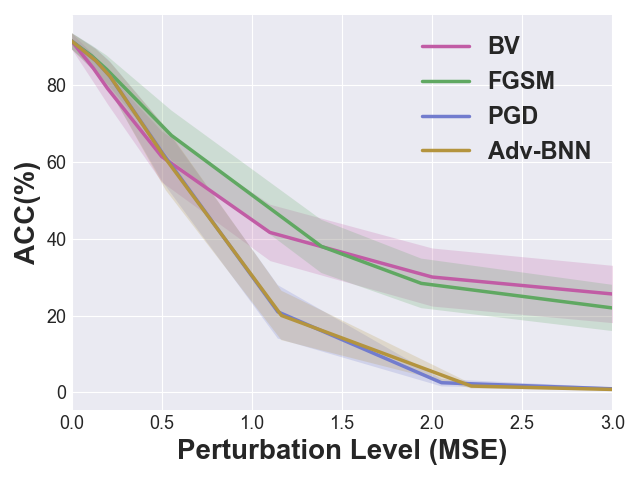}&
		\includegraphics[width=0.37\textwidth]{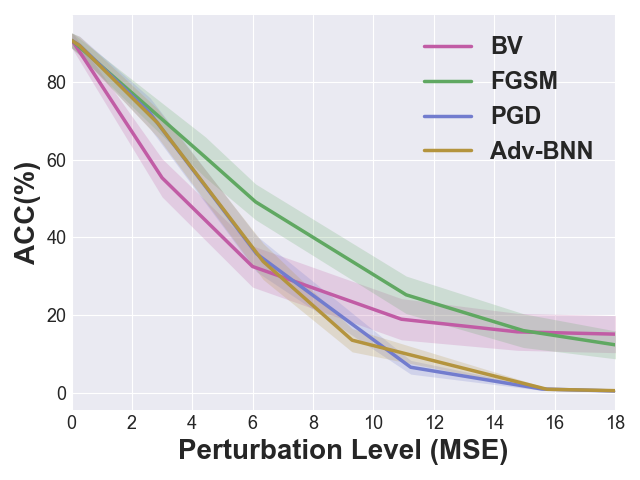}&
		\includegraphics[width=0.37\textwidth]{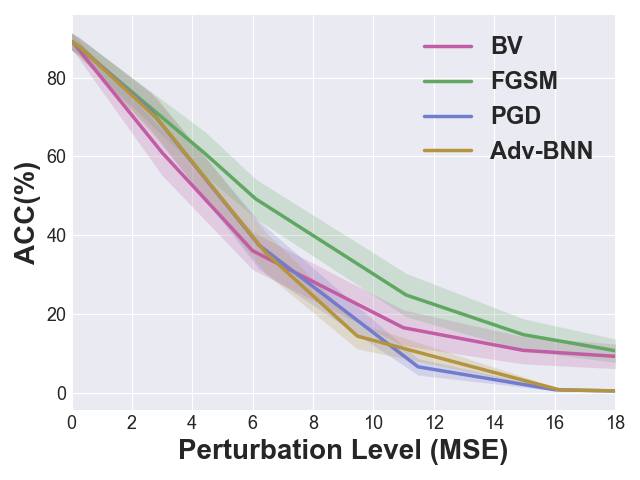}&
		\includegraphics[width=0.37\textwidth]{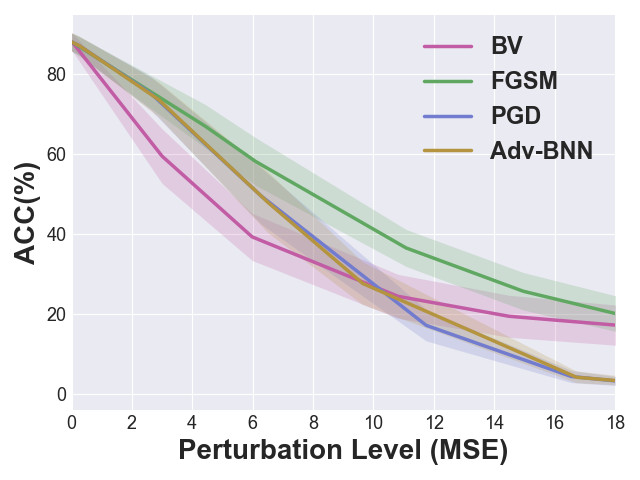}\\
		(a)  ResNet50--No Adversarial Training & (b) ResNet50--FGSM Adv-Training&  (c) ResNet50--BV Adv-Training& (d) ResNet50--PGD Adv-Training
\end{tabular}
\caption{The effect of training with FGSM, PGD, and BV on bias and variance of the accuracy on CIFAR-10 dataset against the competing adversarial perturbations. No Adversarial Training refers to training without using adversarial examples.}
\label{fig:acc-cifar10}
\vspace{-0.5cm}
\end{figure*}


\newpage

{\small
\bibliographystyle{plain}
 \bibliography{egbib}
}
\end{document}